\documentclass[10pt,journal, compsoc]{IEEEtran}
\ifCLASSOPTIONcompsoc
\usepackage[nocompress]{cite}
\usepackage{amsmath,graphicx,mathrsfs,booktabs,type1cm,amssymb,threeparttable,multirow, booktabs, changepage, subfigure}
\else
\usepackage{cite}
\usepackage{amsmath,graphicx,mathrsfs,booktabs,type1cm,amssymb,threeparttable,multirow, booktabs, changepage, subfigure}
\fi
\usepackage{footnote}
\usepackage{color}
\usepackage{verbatim}
\usepackage{bm}
\usepackage{makecell, rotating}
\usepackage{enumerate}
\usepackage{amsthm}
\newtheorem{assumption}{Assumption}
\newtheorem{definition}{Definition}
\newtheorem{theorem}{Theorem}
\newtheorem{proof_mine}{Proof of Theorem}
\usepackage[ruled, linesnumbered]{algorithm2e}
\usepackage[colorlinks,linkcolor=blue]{hyperref}
\usepackage[table]{xcolor}

\ifCLASSINFOpdf
\else
\fi

\begin{document}
	
	\title{IRNet: Iterative Refinement Network for \\ Noisy Partial Label Learning}

	\author{Zheng~Lian,~\IEEEmembership{Senior Member},
		Mingyu~Xu,~\IEEEmembership{}
		Lan~Chen,~\IEEEmembership{}
		Licai~Sun,~\IEEEmembership{} 
		Bin~Liu,~\IEEEmembership{}
		Lei~Feng,~\IEEEmembership{} \\
		and~Jianhua~Tao,\IEEEmembership{Senior Member} 

		\IEEEcompsocitemizethanks{
			
			\IEEEcompsocthanksitem Zheng Lian, Lan Chen, and Bin Liu are with the National Key Laboratory for Multi-modal Artificial Intelligence Systems, Institute of Automation, Chinese Academy of Sciences, Beijing, China.
			E-mail: lianzheng2016@ia.ac.cn; chenlan2016@ia.ac.cn; liubin@nlpr.ia.ac.cn.
			\protect

            \IEEEcompsocthanksitem Mingyu Xu is with the seed group of ByteDance, China.
			E-mail: xumingyu.516@bytedance.com.
			\protect
			
			\IEEEcompsocthanksitem Licai Sun is with the University of Oulu, Finland.
			E-mail: licai.sun@oulu.fi.
			\protect
			
			\IEEEcompsocthanksitem Lei Feng is with the School of Computer Science and Engineering, Southeast University, Nanjing, China.
			E-mail: fenglei@seu.edu.cn.
			\protect
			
			\IEEEcompsocthanksitem Jianhua Tao is with the Department of Automation, Tsinghua University, Beijing, China, and Beijing National Research Center for Information Science and Technology, Tsinghua University, Beijing, China.
			E-mail: jhtao@tsinghua.edu.cn.
			\protect
		}
		
		\thanks{Manuscript received \(\rm{xxxxxxxx}\); revised \(\rm{xxxxxxxx}\). (Corresponding authors: Jianhua Tao, Lei Feng)
		}
	}

	\markboth{IEEE TRANSACTIONS ON PATTERN ANALYSIS AND MACHINE INTELLIGENCE}%
	{Shell \MakeLowercase{\textit{et al.}}: Bare Demo of IEEEtran.cls for Computer Society Journals}

	\IEEEtitleabstractindextext{%
		\begin{abstract}
			Partial label learning (PLL) is a typical weakly supervised learning, where each sample is associated with a set of candidate labels. Its basic assumption is that the ground-truth label must be in the candidate set, but this assumption may not be satisfied due to the unprofessional judgment of annotators. Therefore, we relax this assumption and focus on a more general task, noisy PLL, where the ground-truth label may not exist in the candidate set. To address this challenging task, we propose a novel framework called ``Iterative Refinement Network (IRNet)'', aiming to purify noisy samples through two key modules (i.e., noisy sample detection and label correction). To achieve better performance, we exploit smoothness constraints to reduce prediction errors in these modules. Through theoretical analysis, we prove that IRNet is able to reduce the noise level of the dataset and eventually approximate the Bayes optimal classifier. Meanwhile, IRNet is a plug-in strategy that can be integrated with existing PLL approaches. Experimental results on multiple benchmark datasets show that IRNet outperforms state-of-the-art approaches on noisy PLL. \textcolor[rgb]{0.93,0.0,0.47}{Our source code is available at: https://github.com/zeroQiaoba/IRNet.}
		\end{abstract}
		
		}
	
	\maketitle
	\IEEEdisplaynontitleabstractindextext
	\IEEEpeerreviewmaketitle
	\IEEEraisesectionheading{\section{Introduction}\label{sec:introduction}}
	\IEEEPARstart{P}{artial} label learning (PLL) \cite{yan2020partial, zhang2021exploiting} (also called ambiguous label learning \cite{chen2014ambiguously, chen2017learning} and superset label learning \cite{liu2014learnability, hullermeier2015superset}) is a specific type of weakly supervised learning \cite{zhou2018brief}. In PLL, each sample is associated with a set of candidate labels, only one of which is the ground-truth label. Due to the high monetary cost of accurately labeled data, PLL has become an active research area in many tasks, such as web mining \cite{huiskes2008mir}, object annotation \cite{liu2012conditional, chen2017learning}, and ecological informatics \cite{briggs2012rank}.
	
	\begin{figure}[t]
		\begin{center}
			
			\subfigure[crowdsourced annotation]{
				\label{Figure1-1}
				\centering
				\includegraphics[width=0.43\linewidth]{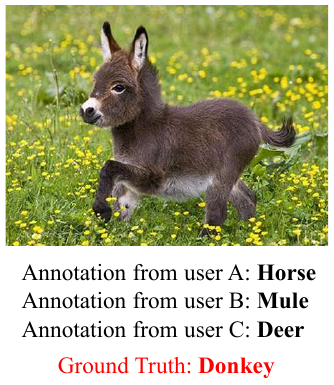}
			} 
			\subfigure[automatic face naming]{
				\label{Figure1-2}
				\centering
				\includegraphics[width=0.476\linewidth]{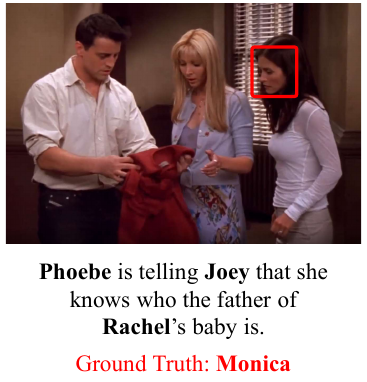}
			}
			
		\end{center}
		\caption{Typical applications of noisy PLL. (a) Candidate labels can be provided by crowdsourcing. However, the ground-truth label may not be in the candidate set. (b) Candidate names can be extracted from the text caption. However, there are general cases of faces without names.}
		\label{Figure1}
	\end{figure}

	Unlike supervised learning \cite{caruana2006empirical}, the ground-truth label is hidden in the candidate set and invisible to PLL \cite{feng2018leveraging, xu2021instance}, which increases the difficulty of model training. Researchers have proposed various approaches to address this problem. These methods can be roughly divided into average-based \cite{cour2009learning, hullermeier2006learning} and identification-based methods \cite{wu2022revisiting, xiaambiguity}. In average-based methods, each candidate label has the same probability of being the ground-truth label. They are easy to implement but may be affected by false positive labels \cite{jin2002learning,wang2022partial}. To this end, researchers introduce identification-based methods that treat the ground-truth label as a latent variable and maximize its estimated probability by the maximum margin criterion \cite{nguyen2008classification, yu2016maximum} or the maximum likelihood criterion \cite{jin2002learning,liu2012conditional}. Due to their promising results, identification-based methods have attracted increasing attention recently.
	
	The above PLL methods rely on a fundamental assumption that the ground-truth label must be in the candidate set. However, this assumption may not be satisfied in real-world scenarios \cite{cid2012proper, lv2023robustness}. Fig. \ref{Figure1} shows some typical examples. In crowdsourced annotations (see Fig. \ref{Figure1-1}), different annotators assign distinct labels to the same image. However, due to the complexity of the image and the unprofessional judgment of the annotator, the ground-truth label may not be in the candidate set. Another typical application is automatic face naming (see Fig. \ref{Figure1-2}). An image with faces is often associated with the text caption, by which we can roughly know who is in the image. However, it is hard to guarantee that all faces have corresponding names in the text caption. Therefore, we relax the assumption of PLL and focus on a more general problem, noisy PLL, where the ground-truth label may not exist in the candidate set. Due to its intractability, few works have studied this problem.

	The core challenge of noisy PLL is how to cope with noisy samples while fully utilizing the provided candidate set. In this paper, we propose a novel framework called ``Iterative Refinement Network (IRNet)''. It is a multi-round technique designed to purify noisy samples through two key modules, i.e., noisy sample detection and label correction. To guarantee the performance of these modules, we further exploit smoothness constraints (i.e., the model's output should not be significantly affected by small input changes) \cite{van2020survey} to increase the reliability of prediction results. Ideally, we can approximate the performance of the traditional PLL if all noisy samples are purified. We also perform theoretical analysis and prove the feasibility of our method. IRNet is a plug-in framework that can be easily integrated with existing PLL methods. Qualitative and quantitative results demonstrate that IRNet outperforms currently advanced approaches under noisy conditions. The main contribution of this paper can be summarized as follows:
	\begin{itemize}
		\item Unlike the traditional PLL where the ground-truth label must be in the candidate set, we focus on a more general task (i.e., noisy PLL) and propose a novel framework (i.e., IRNet) to solve this task.
		
		\item IRNet is a plug-in technique with strong motivation and theoretical guarantees. We also systematically analyze the importance of each component in IRNet.
			
		\item Experimental results on benchmark datasets show the effectiveness of our method. IRNet is superior to existing state-of-the-art methods on noisy PLL.
	\end{itemize}
	
	The remainder of this paper is organized as follows: In Section \ref{sec2}, we briefly review some recent works. In Section \ref{sec3}, we propose a novel framework for noisy PLL with theoretical guarantees. In Section \ref{sec4}, we introduce our experimental datasets, comparison systems, and implementation details. In Section \ref{sec5}, we conduct experiments to demonstrate the effectiveness of our method. In Section~\ref{sec7}, we provide an in-depth discussion and comparison across different tasks. Finally, we conclude this paper and discuss our future work in Section \ref{sec6}.

	\section{Related Works}
	\label{sec2}
	
	\subsection{Non-deep Partial Label Learning}
	In PLL, the ground-truth label is concealed in the candidate set. Researchers have proposed various non-deep PLL methods to disambiguate candidate labels. We roughly divide them into average-based and identification-based methods.
	
	\textbf{Average-based methods.} They assume that each candidate label has an equal probability of being the ground-truth label. For parametric models, Cour et al. \cite{cour2009learning, cour2011learning} proposed to distinguish the average output of candidate and non-candidate labels. For instance-based models, Hullermeier et al. \cite{hullermeier2006learning} estimated the ground-truth label of each sample by voting on the label information of its neighborhood. However, these average-based methods can be severely affected by false positive labels in the candidate set \cite{jin2002learning, wang2022partial}.
	
	\textbf{Identification-based methods.} They address the shortcomings of average-based methods by directly identifying the ground-truth label and maximizing its estimated probability \cite{nguyen2008classification}. To implement these methods, researchers have exploited various techniques, such as maximum margin \cite{nguyen2008classification, yu2016maximum} and maximum likelihood criteria \cite{jin2002learning,liu2012conditional}. For example, Nguyen et al. \cite{nguyen2008classification} maximized the margin between the maximum output of candidate and non-candidate labels. However, it cannot distinguish the ground-truth label from other candidate labels. To this end, Yu et al. \cite{yu2016maximum} directly maximized the margin between the ground-truth label and all other labels. Unlike the maximum margin criterion, Jin et al. \cite{jin2002learning} iteratively optimized the latent ground-truth label and trainable parameters via the maximum likelihood criterion. Considering the fact that different samples and distinct candidate labels should contribute differently to the learning process, Tang et al. \cite{tang2017confidence} further utilized the weight of each sample and the confidence of each candidate label to facilitate disambiguation. They exploited the model outputs to estimate the confidence of candidate labels. Differently, Zhang et al. \cite{zhang2015solving, zhang2016partial} and Gong et al. \cite{gong2017regularization} performed confidence estimation based on the assumption that samples that were close in feature space tended to share the same label. But limited by linear models, these non-deep PLL methods are usually optimized in an inefficient manner \cite{yao2020deep}.
	
	\subsection{Deep Partial Label Learning}
	Deep learning greatly promotes the development of PLL. To implement deep PLL, researchers have proposed some training objectives compatible with stochastic optimization. For example, Lv et al. \cite{lv2020progressive} introduced a self-training strategy that exploited model outputs to disambiguate candidate labels. Feng et al. \cite{feng2020provably} assumed that each incorrect label had a uniform probability of being the candidate labels. Based on this assumption, they further proposed classifier- and risk-consistent algorithms with theoretical guarantees. Wen et al. \cite{wen2021leveraged} relaxed the data generation assumption in \cite{feng2020provably}. They introduced a family of loss functions that considered the trade-off between losses on candidate and non-candidate labels. Recently, Wang et al. \cite{wang2022pico} proposed a contrastive-learning-based framework that exploited the prototype-based disambiguation algorithm to identify the ground-truth label from the candidate set. Wu et al. \cite{wu2022revisiting} performed consistency regularization on candidate labels and employed supervised learning on non-candidate labels, achieving promising results under varying ambiguity levels.
	
	The above PLL methods are based on a fundamental assumption that the ground-truth label of each sample must be in the candidate set \cite{lv2020progressive, feng2020provably}. But due to the unprofessional judgment of the annotators, this assumption may not be satisfied in real-world scenarios.
	
	\subsection{Noisy Partial Label Learning}
	More recently, some researchers have noticed the importance of noisy PLL. Unlike traditional PLL, it considers a scenario where the ground-truth label may not be in the candidate set \cite{cid2012proper, lv2023robustness}. Typically, Lv et al. \cite{lv2023robustness} utilized the noise-robust loss functions to avoid overemphasizing noisy samples, which also caused them to fail to use the useful information in these samples. Differently, IRNet can purify noisy samples, allowing us to leverage them to learn more discriminative classifiers. Wang et al. \cite{wang2023pico+} relied on distance-based noisy sample detection and learned robust classifiers by semi-supervised learning. Differently, IRNet detects noisy samples and corrects labels through distinct strategies, with strong motivations and theoretical guarantees.
	
	\subsection{Learning from Noisy Labels}
	How to cope with incorrect training samples is one of the most important issues in machine learning \cite{han2020survey, song2022learning, liu2020early}. A common approach is to design noise-robust loss functions. For example, Zhang et al. \cite{zhang2018generalized} introduced Generalized Cross Entropy (GCE), which combined the noise-robust but poor-performing Mean Absolute Error (MAE) with the noise-sensitive but good-performing Categorical Cross Entropy (CCE). Wang et al. \cite{wang2019symmetric} proposed Symmetric Cross Entropy (SCE) that boosted CCE with a noise-robust counterpart Reverse Cross Entropy (RCE). All these methods extend the commonly used CCE to address its overfitting problem in the presence of noisy labels.
	
	Recently, a promising way of handling noisy labels is to separate clean and noisy samples. Typically, Jiang et al. \cite{jiang2018mentornet} considered samples with smaller losses as clean ones. They selected clean samples to guide the training of the main network. However, this method is similar to self-training and may suffer from the confirmation bias problem \cite{tarvainen2017mean}. To address this issue, Han et al. \cite{han2018co} maintained two branches. Each branch selected clean samples through the small-loss criterion to train the other branch. Yu et al. \cite{yu2019does} further improved the diversity of two branches by training on disagreement data. Different from the aforementioned methods, Li et al. \cite{li2020dividemix} proposed DivideMix, a well-known strategy that treated noisy samples as unlabeled data and trained the model in a semi-supervised learning manner.
	
	It is worth noting that noisy PLL is not a simple combination of PLL and noisy labels. In noisy label learning, previous works attempt to separate clean and noisy samples through small loss tricks \cite{jiang2018mentornet, han2018co}. However, noisy PLL does not provide ground-truth labels for loss calculation, but a set of candidate labels. How to detect noisy samples in this setting is an important but challenging issue. Different methods bring distinct performances. This paper introduces IRNet with strong motivation and theoretical guarantees. Experimental results in Table \ref{Table3} show the effectiveness of our method compared with other noise-robust techniques.

	\begin{table}[t]
		\centering
		\renewcommand\tabcolsep{2.4pt}
		\renewcommand\arraystretch{1.20}
		\caption{Summary of main mathematical notations.}
		\label{Table1}
		\begin{tabular}{c|c}
			\hline
			Notations & Mathematical Meanings \\
			\hline
			\hline
			$\mathcal{X}, \mathcal{Y}$ & feature and label space \\
			$N, C$ & number of samples and labels \\
			$\mathcal{D}_S, \tilde{\mathcal{D}}_S$ & partially and fully labeled datasets \\
			$\tilde{\mathcal{D}}_S^c, \tilde{\mathcal{D}}_S^n$ & clean subset and noisy subset of $\tilde{\mathcal{D}}_S$ \\
			$q, \eta$ & ambiguity level and noise level \\
			$x,y,\bar{y}$ & feature, ground-truth label and incorrect label \\
			$y_j(x)$ & the $j^{th}$ element of one-hot encoded label of $x$ \\
			$S(x)$ & mapping function from $x$ to its candidate set \\
			$\omega_{1}, \omega_{2}$ & indicator for clean and noisy samples \\
			$\tau,\tau_\epsilon$ & feature and boundary for noisy sample detection \\
			$\phi_c, \phi_n$ & hit accuracy for $\tilde{\mathcal{D}}_S^c$ and $\tilde{\mathcal{D}}_S^n$ \\
			$E_{\max}$ & maximum number of epochs \\
                $T_{\max}$ & maximum number of iteration in each epoch \\
			$e_0$ & epoch to start correction \\
			$\mathcal{A}(x), K$ & $\mathcal{A}(x)$ contains $K$ augmented samples for $x$ \\
			$h(x), h^*(x)$ & any classify and Bayes optimal classifier \\
			$p(y=j|x)$ & posterior probability of $x$ on the label $j$\\
			$y^{x}$ & label with the highest posterior probability \\
			$o^{x}$ & label with the second highest posterior probability \\
			$f_j(x)$ & estimated probability of $x$ on the label $j$ \\
			$u(x),d(u)$ & confidence of $x$ and its density function \\
			$c_*,c^*,l$ & lower bound, upper bound and ratio of $d(u)$ \\
			$L(m)$ & pure level set with the boundary $m$ \\
			$\alpha,\epsilon$ & two values that control the approximate gap \\
			$\lambda_c,\lambda_r,\lambda_g$ & hyper-parameters for noise-robust loss functions \\
			\hline
			
		\end{tabular}
	\end{table}

	\section{Methodology}
	\label{sec3}
\begin{figure*}[t]
		\centering
		\includegraphics[width=\linewidth]{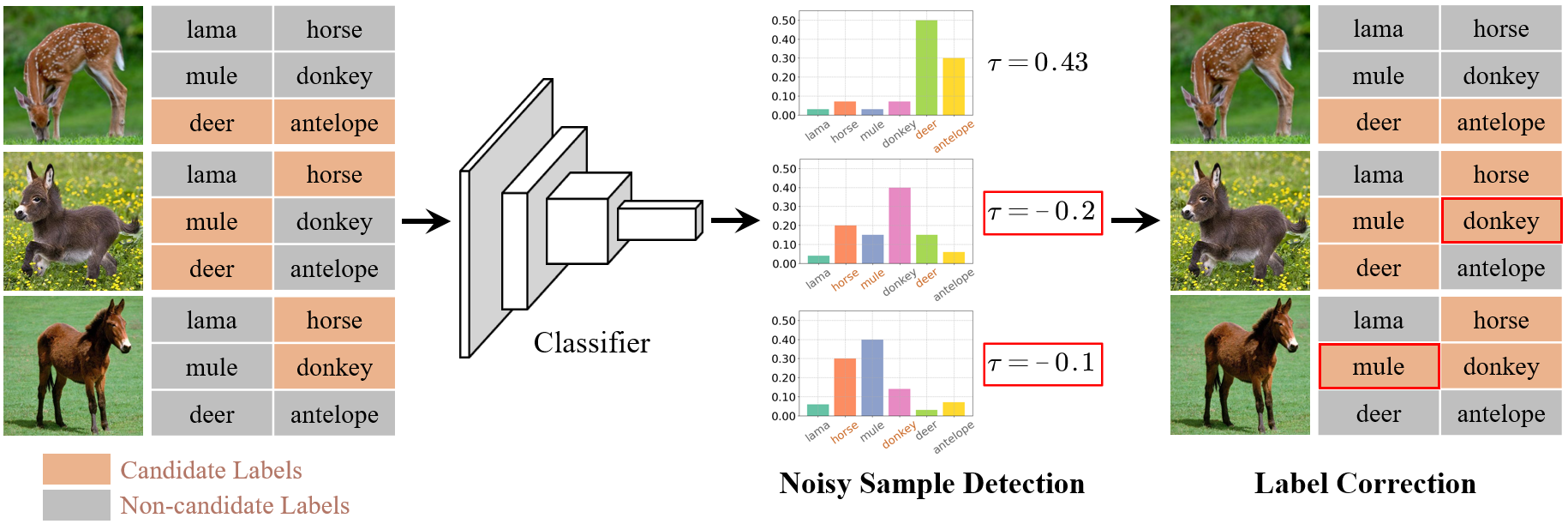}
		\caption{IRNet contains two core modules: noisy sample detection and label correction. First, we use the metric $\tau$ (see Eq. \ref{eq-tau}) to identify noisy samples. Then, we use the non-candidate label with the highest probability to update the candidate set for the detected noisy samples.}
		\label{Figure10}
	\end{figure*}

	In this section, we first formalize the problem statement for noisy PLL. Then, we discuss the motivation and introduce our proposed method in detail. Finally, we conduct theoretical analyze and prove the feasibility of IRNet. Fig. \ref{Figure10} shows the core structure of our method.
		
	\subsection{Problem Definition}
	\label{sec:problem_definition}
	Let $\mathcal{X}$ be the input space and $\mathcal{Y}=\left\{1, 2, \cdots, C\right\}$ be the label space with $C$ distinct classes. We consider a partially labeled dataset $\mathcal{D}_{S}=\{ \left(x_{i}, S(x_{i})\right)\}_{i=1}^{N}$ where the function $S(\cdot)$ maps each sample $x \in \mathcal{X}$ into its corresponding candidate set $S(x) \subseteq \mathcal{Y}$. The goal of PLL is to learn a multi-class classifier $f:\mathcal{X}\rightarrow\mathcal{Y}$ that minimizes the classification risk on the dataset $\mathcal{D}_{S}$. In PLL, the basic assumption is that the ground-truth label $y$ must be in the candidate set $S(x)$. In this paper, we relax this assumption and focus on noisy PLL.
	
	We first introduce some necessary notations. The samples that satisfy $y \in S(x)$ are called \emph{clean} samples; otherwise are called \emph{noisy} samples. We adopt the same generating procedure as previous works to synthesize the candidate set \cite{lv2020progressive, wen2021leveraged}. For each sample, any incorrect label $\bar{y} \in \mathcal{Y} \backslash \{y\}$ has a probability $q$ of being an element in the candidate set. After that, each sample has a probability $\eta$ of being a noisy sample, i.e., $P\left(y \notin S(x)\right)=\eta$. In this paper, we assume that the label space is known and fixed. Both clean and noisy samples have their ground-truth labels in this space. The case where $x$ may be an out-of-distribution sample (i.e., $y \notin \mathcal{Y}$) is left for our future work \cite{hendrycks2017baseline, liang2018enhancing}.
	
	\subsection{Observation and Motivation}
	\label{sec:observation_and_motivation}
	How to effectively deal with noisy samples is an important issue in noisy PLL. For each noisy sample, its ground-truth label $y$ is concealed in the non-candidate set $\{j|j\in\mathcal{Y},j \notin S(x)\}$. A heuristic solution is to purify the noisy sample by moving $y$ from the non-candidate set to the candidate set. To this end, we need to realize two core functions: (1) detect noisy samples in the entire dataset; (2) identify the ground-truth label of each noisy sample for label correction. In this section, we conduct pilot experiments on CIFAR-10 ($q=0.3, \eta=0.3$) \cite{krizhevsky2009learning} and discuss our motivation in detail. 
	
	In the pilot experiments, we consider a fully labeled dataset $\tilde{\mathcal{D}}_S=\{ (x_{i}, y_{i}, S(x_{i}))\}_{i=1}^{N}$ where the ground-truth label $y_{i}$ is associated with each sample $x_{i}$. Suppose $\omega(x_i) \in \{\omega_{1}, \omega_{2}\}$ indicates whether $x_{i}$ is clean or noisy:
	\begin{equation}
	\omega(x_i)=\begin{cases}
	\omega_{1}, y_{i} \in S(x_{i}) \\
	\omega_{2}, y_{i} \notin S(x_{i}).
	\end{cases}
	\end{equation}
	Based on $\omega(x_i)$, we split the dataset $\tilde{\mathcal{D}}_S$ into the clean subset $\tilde{\mathcal{D}}_S^c$ and the noisy subset $\tilde{\mathcal{D}}_S^n$. In this section, $|\tilde{\mathcal{D}}_S^c|$ and $|\tilde{\mathcal{D}}_S^n|$ denote the number of samples in corresponding subsets:
	\begin{equation}
	\tilde{\mathcal{D}}_S^c=\{ \left(x_{i}, y_{i}, S(x_{i})\right) | \omega(x_i)=\omega_{1}, 1 \leqslant i \leqslant N\},
	\end{equation}
	\begin{equation}
	\tilde{\mathcal{D}}_S^n=\{ \left(x_{i}, y_{i}, S(x_{i})\right) | \omega(x_i)=\omega_{2}, 1 \leqslant i \leqslant N\}.
	\end{equation}

	\begin{figure*}[t]
		\begin{center}
			
			\subfigure[Bayes error rate]{
				\label{Figure2-1}
				\centering
				\includegraphics[width=0.3\linewidth]{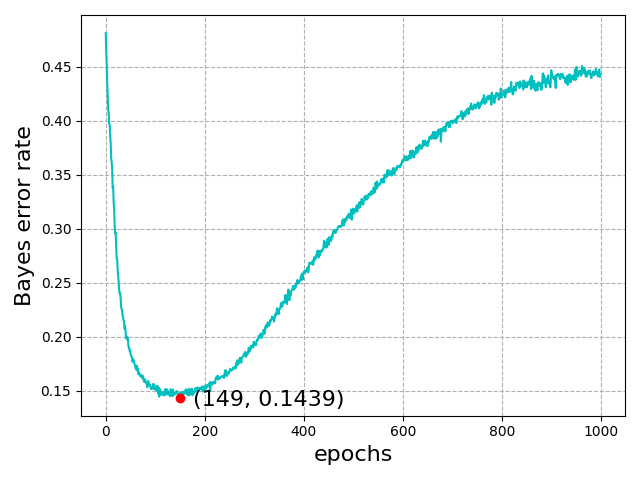}
			} 
			\subfigure[hit accuracy]{
				\label{Figure2-2}
				\centering
				\includegraphics[width=0.3\linewidth]{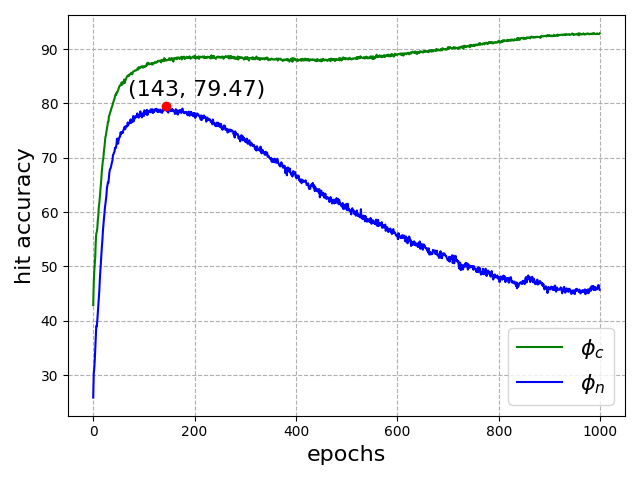}
			}
			\subfigure[validation accuracy]{
				\label{Figure2-3}
				\centering
				\includegraphics[width=0.3\linewidth]{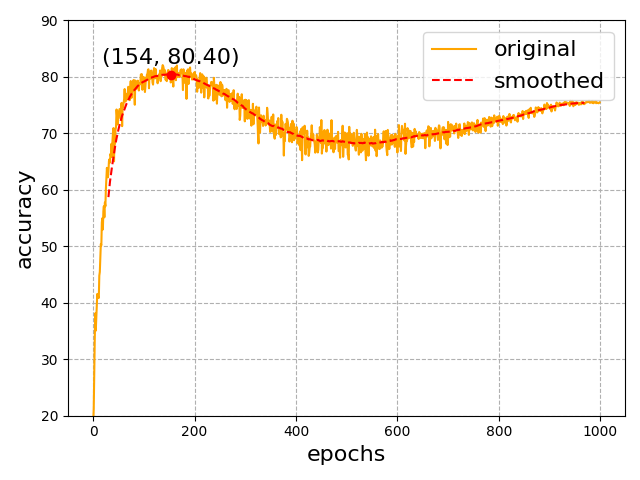}
			}

		\end{center}
		\caption{Visualization of the Bayes error rate, hit accuracy, and validation accuracy on CIFAR-10 ($q=0.3, \eta=0.3$). We mark the minima in (a), the maxima in (b), and the local maxima in (c) with red dots.}
		\label{Figure2}
	\end{figure*}
	
	\subsubsection{Noisy Sample Detection}
	To achieve better performance on unseen data, clean samples usually contribute more than noisy samples during training \cite{jiang2018mentornet}. Therefore, we conjecture that any metric reflecting sample contributions to the learning process can be used for noisy sample detection. In this paper, we focus on the predictive difference between the maximum output of candidate and non-candidate labels \cite{yu2016maximum}, a popular metric in traditional PLL \cite{tang2017confidence}. This metric is calculated as follows:
	\begin{equation}
	\label{eq-tau}
	\tau = \underset{j \in S\left(x\right)}{\max}{f_j\left(x\right)} - \underset{j \notin S\left(x\right)}{\max}{f_j\left(x\right)},
	\end{equation}
	where $f_j\left(x\right)$ represents the estimated probability of the sample $x$ on the label $j$. Here, $\sum_{j=1}^{C}{f_j\left(x\right)=1}$.
	
	In the pilot experiment, we attempt to verify the performance of the feature $\tau$ on noisy sample detection. This value depends on the estimated probabilities and therefore changes every epoch. To evaluate its performance, we need to train additional detectors at each epoch, which is time-consuming and computationally expensive. For convenience, we use the Bayes error rate \cite{hart2000pattern}, which is easy to implement and does not require an additional training process. The calculation formula is shown as follows:
	\begin{equation}
	P(error|\tau) = \begin{cases}
	P(\omega_2|\tau), P(\omega_1|\tau) > P(\omega_2|\tau) \\
	P(\omega_1|\tau), P(\omega_1|\tau) < P(\omega_2|\tau),
	\end{cases}
	\end{equation}
	\begin{equation}
	\begin{split}
	P(error) &= \int{P(error|\tau)p(\tau)d\tau} \\
	&= \int\limits_{\mathcal{R}_1}{P(\omega_2|\tau)p(\tau)d\tau} + \int\limits_{\mathcal{R}_2}{P(\omega_1|\tau)p(\tau)d\tau} \\
	&= \int\limits_{\mathcal{R}_1}{p(\tau|\omega_2)P(\omega_2)d\tau} + \int\limits_{\mathcal{R}_2}{p(\tau|\omega_1)P(\omega_1)d\tau}.
	\end{split}
	\end{equation}
	Based on whether satisfying $P(\omega_1|\tau) > P(\omega_2|\tau)$, we divide the space into two regions $\mathcal{R}_1$ and $\mathcal{R}_2$. Here, $P(\omega_1)$ and $P(\omega_2)$ are the prior probabilities for noisy sample detection. Since we cannot obtain such priors, we set them to $P(\omega_1)=P(\omega_2)=0.5$. Meanwhile, we leverage histograms to approximate the class-conditional probability density functions $P(\tau|\omega_1)$ and $P(\tau|\omega_2)$.
	
	Experimental results are presented in Fig. \ref{Figure2-1}. A lower Bayes error rate means better performance. If we randomly determine whether a sample is clean or noisy, the Bayes error rate is $P(error)=0.50$. From Fig. \ref{Figure2-1}, we observe that the Bayes error rate can reach $P(error)=0.14$, much lower than random guessing. Therefore, we can infer from $\tau$ whether a sample is more likely to be clean or noisy.
	
	\subsubsection{Label Correction}
	After noisy sample detection, we try to identify the ground-truth label of each noisy sample so that we can correct it by moving the predicted label into its candidate set. For the clean sample, previous works usually select the candidate label with the highest probability $\arg\max_{j \in S(x)}{f_j\left(x\right)}$ as the predicted label \cite{tang2017confidence, xu2019partial}. For the noisy sample, the ground-truth label is concealed in the non-candidate set. Naturally, we conjecture that the non-candidate label with the highest probability $\arg\max_{j \notin S(x)}{f_j\left(x\right)}$ can be regarded as the predicted label for the noisy sample. To verify this hypothesis, we first define an evaluation metric called \emph{hit accuracy}. The calculation formula is shown as follows:
	\begin{equation}
	\phi_c = \frac{1}{\left|\tilde{\mathcal{D}}_S^c\right|}{\sum\nolimits_{\left(x,y,S\left(x\right)\right) \in \tilde{\mathcal{D}}_S^c}}{\mathbb{I}\left(y = \underset{j \in S(x)}{\arg\max}{f_j(x)}\right)},
	\end{equation}
	\begin{equation}
	\phi_n = \frac{1}{\left|\tilde{\mathcal{D}}_S^n\right|}{\sum\nolimits_{\left(x,y,S\left(x\right)\right) \in \tilde{\mathcal{D}}_S^n}}{\mathbb{I}\left(y = \underset{j \notin S(x)}{\arg\max}{f_j(x)}\right)},
	\end{equation}
	where $\mathbb{I}\left(\cdot\right)$ is an indicator function. $\phi_c$ is the hit accuracy for the clean subset $\tilde{\mathcal{D}}_S^c$ and $\phi_n$ is the hit accuracy for the noisy subset $\tilde{\mathcal{D}}_S^n$. This evaluation metric measures the performance of our method on ground-truth label detection. Higher hit accuracy means better performance.
	
	Experimental results are shown in Fig. \ref{Figure2-2}. If we randomly select a non-candidate label as the predicted label for the noisy sample, $\phi_n$ is calculated as follows:
	\begin{equation}
	\begin{split}
	\phi_n &= \frac{1}{C-\mathbb{E}_{x \sim \tilde{\mathcal{D}}_S^n}\left[\left|S(x)\right|\right]} \\
	&= \frac{1}{C-\left(1+ \left(C-1\right) * q\right)},
	\end{split}
	\end{equation}
	where $\mathbb{E}_{x \sim \tilde{\mathcal{D}}_S^n}\left[\left|S(x)\right|\right]$ denotes the expectation on the number of candidate labels. Since we conduct pilot experiments on CIFAR-10 ($q=0.3, \eta=0.3$), the hit accuracy of random guessing is $\phi_n=15.87$. From Fig. \ref{Figure2-2}, we observe that our method can reach $\phi_n=79.47$, much higher than random guessing. These results demonstrate the effectiveness of our method on ground-truth label detection for noisy samples.

	\begin{figure*}[t]
		\begin{center}
			\subfigure[100 epochs, $P(error)$=0.15]{
				\label{Figure3-1}
				\centering
				\includegraphics[width=0.23\linewidth, trim=30 0 0 0]{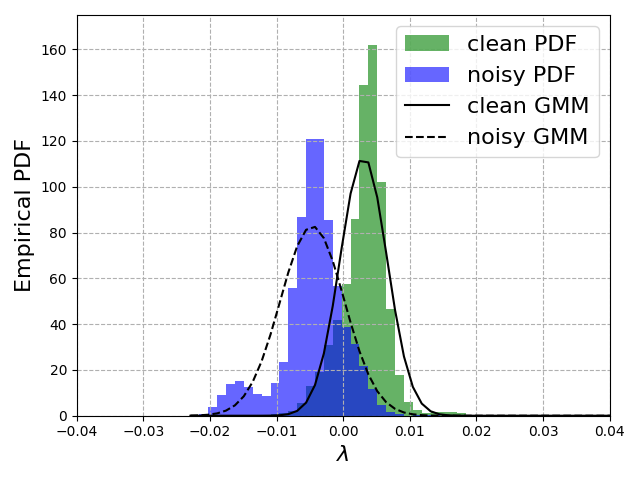}
			} 
			\subfigure[149 epochs, $P(error)$=0.14]{
				\label{Figure3-2}
				\centering
				\includegraphics[width=0.23\linewidth, trim=30 0 0 0]{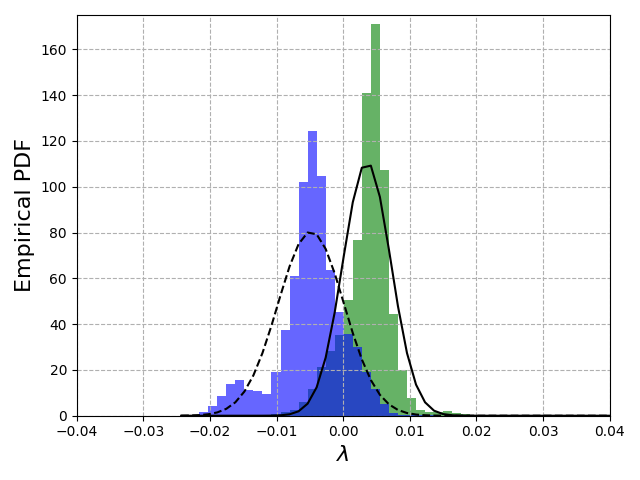}
			}
			\subfigure[200 epochs, $P(error)$=0.15]{
				\label{Figure3-3}
				\centering
				\includegraphics[width=0.23\linewidth, trim=30 0 0 0]{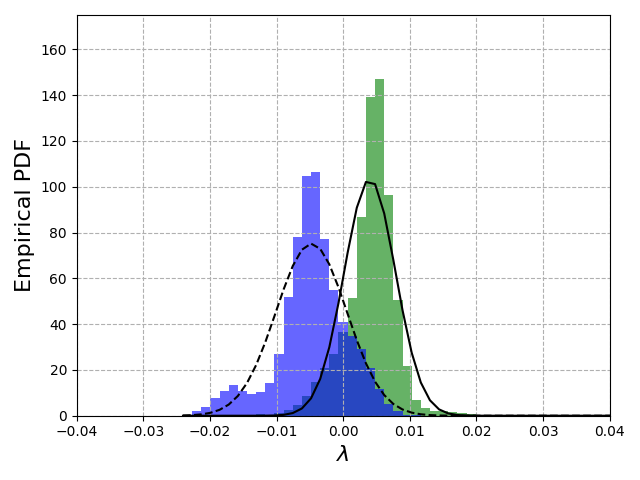}
			} 
			\subfigure[300 epochs, $P(error)$=0.19]{
				\label{Figure3-4}
				\centering
				\includegraphics[width=0.23\linewidth, trim=30 0 0 0]{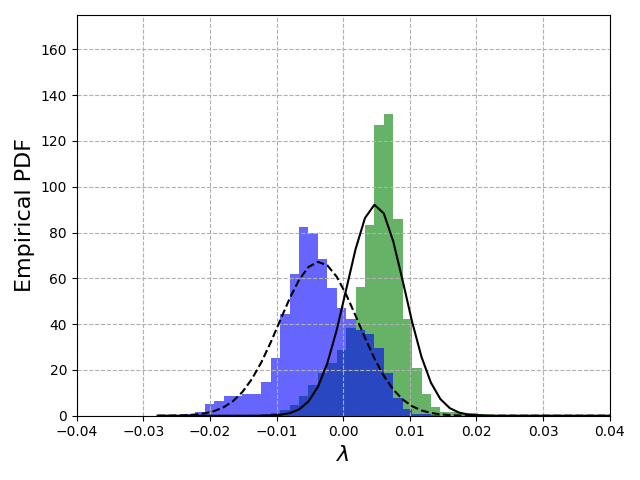}
			}
			
			\subfigure[400 epochs, $P(error)$=0.26]{
				\label{Figure3-5}
				\centering
				\includegraphics[width=0.23\linewidth, trim=30 0 0 0]{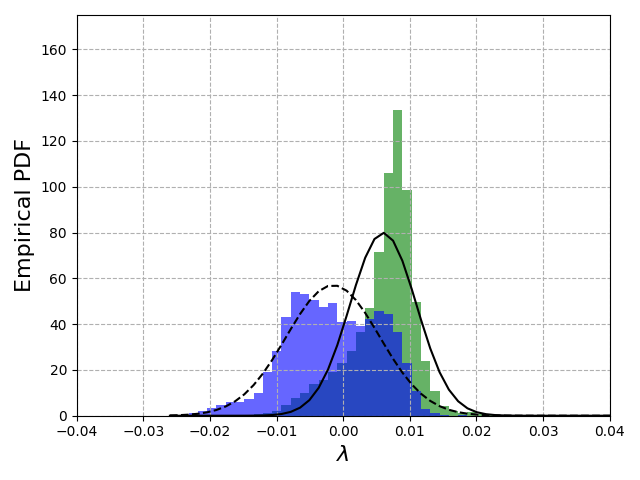}
			} 
			\subfigure[500 epochs, $P(error)$=0.32]{
				\label{Figure3-6}
				\centering
				\includegraphics[width=0.23\linewidth, trim=30 0 0 0]{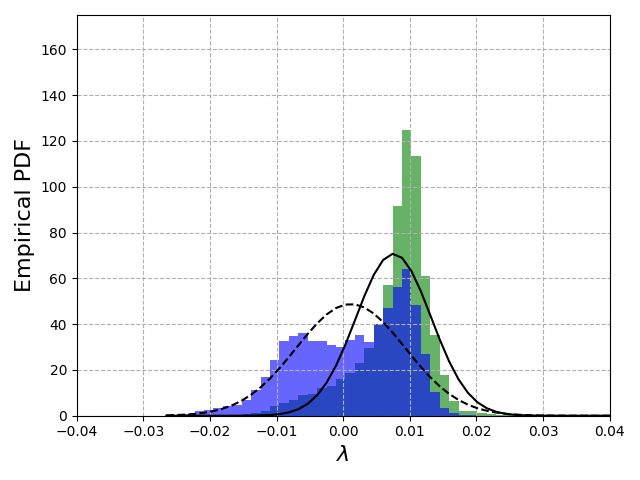}
			}
			\subfigure[600 epochs, $P(error)$=0.36]{
				\label{Figure3-7}
				\centering
				\includegraphics[width=0.23\linewidth, trim=30 0 0 0]{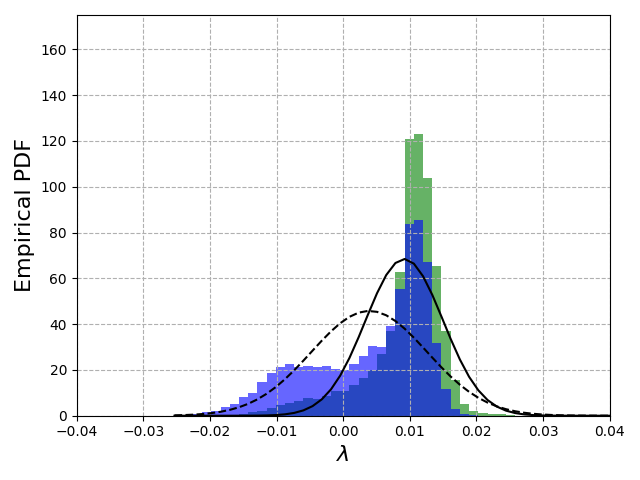}
			} 
			\subfigure[700 epochs, $P(error)$=0.40]{
				\label{Figure3-8}
				\centering
				\includegraphics[width=0.23\linewidth, trim=30 0 0 0]{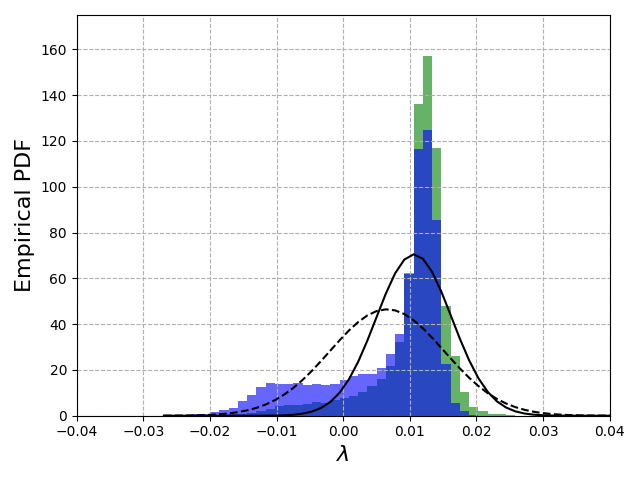}
			}
			
		\end{center}
		\caption{Empirical PDF and estimated GMM models on CIFAR-10 ($q=0.3, \eta=0.3$) with increasing training epochs.}
		\label{Figure3}
	\end{figure*}

	\subsection{Iterative Refinement Network (IRNet)}
	\label{sec:3.3}
	Motivated by the above observations, we propose a simple yet effective framework for noisy PLL. Our method consists of two key modules, i.e., noisy sample detection and label correction. However, we find some challenges during implementation: (1) As shown in Fig. \ref{Figure2-1}$\sim$\ref{Figure2-2}, these modules perform poorly at the early epochs of training. Therefore, we need to find an appropriate epoch $e_{0}$. Before $e_{0}$, we start with a warm-up period with traditional PLL approaches. After $e_{0}$, we exploit IRNet for label correction. In this paper, we refer to $e_{0}$ as \emph{correction epoch}; (2) Unlike pilot experiments in Section \ref{sec:observation_and_motivation}, ground-truth labels are not available in real-world scenarios. Therefore, we need to find suitable unsupervised techniques to realize these modules. In this section, we illustrate our solutions to the above challenges.
	
	\subsubsection{Choice of Correction Epoch}
	An appropriate $e_{0}$ is when both noisy sample detection and label correction can achieve good performance. As shown in Fig. \ref{Figure2-1}$\sim$\ref{Figure2-2}, the performance of these modules first increases and then decreases. Meanwhile, these modules achieve the best performance at close epochs. To find a suitable $e_{0}$, we first reveal the reasons behind these phenomena.
	
	Previous works have demonstrated that when there is a mixture of clean and noisy samples, networks tend to fit the former before the latter \cite{jiang2018mentornet, arazo2019unsupervised}. Based on this theory, in noisy PLL, networks start fitting clean samples so that the performance of these modules first increases. At the late epochs of training, networks tend to fit noisy samples. Since the ground-truth label of the noisy sample is not in the candidate set, networks may assign wrong predictions to the candidate set and gradually become overconfident during training. This process harms the discriminative performance of $\tau$ and leads to a decrement in $\phi_n$. Therefore, the epoch when networks start fitting noisy samples is a suitable $e_{0}$.
	
	Empirically, networks cannot perform accurately on unseen data when fitting noisy samples, resulting in a drop in validation accuracy. Therefore, we conjecture that validation accuracy can be used to guide the model to find an appropriate $e_{0}$. To verify this hypothesis, we visualize the curve of validation accuracy in Fig. \ref{Figure2-3}. We observe that it first increases and then decreases, the same trend as in Fig. \ref{Figure2-1}$\sim$\ref{Figure2-2}. At the late epochs of training, it gradually increases and converges to a stable value. Since $\phi_c$ keeps increasing during training (see Fig. \ref{Figure2-2}), the increment brought by clean samples is greater than the decrement brought by noisy samples, resulting in a slight improvement in validation accuracy. Therefore, we choose the epoch of the first local maximum as $e_{0}$. Experimental results demonstrate that the epoch determined by our method (the $154^{th}$ epoch) is close to the best epoch for noisy sample detection (the $149^{th}$ epoch) and label correction (the $143^{th}$ epoch). These results verify the effectiveness of our selection strategy.

	\subsubsection{Noisy Sample Detection}
	In this section, we attempt to find an appropriate approach for noisy sample detection. Since ground-truth labels are unavailable in real-world scenarios, we rely on unsupervised techniques to realize this function.
	
	Previous works \cite{li2020dividemix} usually exploit the Gaussian Mixture Model (GMM) \cite{permuter2006study}, a popular unsupervised modeling technique. To verify its performance, we show the empirical PDF and estimated GMM on the feature $\tau$ for clean and noisy subsets. From Fig. \ref{Figure3}, we observe that the Gaussian cannot accurately estimate the distribution. Hence, we need to find a more effective strategy to detect noisy samples.
	
	From Fig. \ref{Figure3}, we observe some interesting phenomena. As the number of training iterations increases, the range of $\tau$ is relatively fixed and most samples satisfy $\tau \in \left[-0.03, 0.03\right]$. Meanwhile, samples meeting $\tau \leq -0.008$ have a high probability of being noisy samples. Inspired by the above phenomena, we determine the samples satisfying $\tau \leq -\tau_\epsilon$ as the noisy samples, where $\tau_\epsilon$ is a user-defined parameter.

	\subsubsection{Label Correction}
	After noisy sample detection, we try to identify the ground-truth label of each noisy sample for label correction. Inspired by pilot experiments in Section \ref{sec:observation_and_motivation}, we exploit the non-candidate label with the highest probability as the predicted label for the detected noisy sample. However, prediction results cannot be completely correct. To achieve better performance, we need to further reduce prediction errors.
	
	A straightforward solution is to encourage the model's output not to be significantly affected by natural and small input changes, i.e., the smoothness assumption \cite{van2020survey}. This process can increase the reliability of the prediction results. Specifically, we assume that each sample has a set of augmented versions $\mathcal{A}(x)=\left\{ \text{Aug}_k(x)|1 \leqslant k \leqslant K\right\}$, where $K$ denotes the number of augmentations. $\text{Aug}_k(x)$ represents the $k^{th}$ augmented version for $x$. We select the sample satisfying the following conditions for label correction:
	\begin{equation}
	\label{eq11}
	\underset{j \in S(x)}{\max}{f_j\left(x\right)} - \underset{j \notin S(x)}{\max}{f_j\left(x\right)} \leq -\tau_\epsilon,
	\end{equation}
	\begin{equation}
	\underset{j \in S(x)}{\max}{f_j\left(z\right)} - \underset{j \notin S(x)}{\max}{f_j\left(z\right)} \leq -\tau_\epsilon, \forall z \in \mathcal{A}(x),
	\end{equation}
	\begin{equation}
	\label{eq13}
	\underset{j \notin S(x)}{\arg\max}{f_j\left(x\right)} = \underset{j \notin S(x)}{\arg\max}{f_j\left(z\right)}, \forall z \in \mathcal{A}(x).
	\end{equation}
	Therefore, we rely on stricter criteria for noisy sample detection and label correction. Besides the original sample, its augmented versions should meet additional conditions. Experimental results in Section \ref{sec:augmentation} show that this strategy can reduce prediction errors and improve the classification performance under noisy conditions. The pseudo-code of IRNet is summarized in Algorithm \ref{alg-1}.

	\begin{algorithm*}[t]
		\caption{IRNet Algorithm}
		\label{alg-1}
		\KwIn{PLL training set $\mathcal{D}_S=\{ \left(x_{i}, S(x_{i})\right)\}_{i=1}^{N}$, the predictive model $f$, the margin $\tau_\epsilon$, the number of augmentations $K$, the number of epochs $E_{\max}$, the number of iteration $T_{\max}$. Specifically, $T_{\max}$ depends on the number of training samples and the batch size, given by $T_{\max}=\lceil N/N_b \rceil$.}
		\KwOut{The optimized model $f$.}
		\BlankLine
		
		Initialize model parameters for $f$;
		
		\For{$e=1,\cdots, E_{\max}$}{
			
			Shuffle $\mathcal{D}_S$;
			
			\For{$t=1,\cdots, T_{\max}$}{
				
				Sample mini-batch $\mathcal{B}_t=\{ \left(x_{i}, S(x_{i})\right)\}_{i=1}^{N_b}$;
				
				Train the predictive model $f$ on $\mathcal{B}_t$;
				
			}
			
			Evaluate $f$ using validation accuracy and save the result in $res^{[e]}$;
			
			\If{$e \geqslant 10$}{
				Check the convergence condition $\sum\limits_{i=e-9}^{e}\left(res^{[i]}-res^{[i-1]}\right)<10^{-5}$;
				
				\If{converged}{
					$e_0=e$;
					
					break;
				}
			}
		}
		
		~\\
		
		\For{$e=e_0,\cdots, E_{\max}$}{
			Shuffle $\mathcal{D}_S$;
			
			\For{$t=1,\cdots, T_{\max}$}{
				Sample mini-batch $\mathcal{B}_t=\{ \left(x_{i}, S(x_{i})\right)\}_{i=1}^{N_b}$;
				
				Train the predictive model $f$ on $\mathcal{B}_t$;
				
				\If{$i=1,\cdots, N_b$}{
					\If{$x_i$ satisfies the conditions in Eq. \ref{eq11}$\sim$\ref{eq13}}{
						Identify $x_i$ as the noisy sample;
						
						Gain the predicted label $\arg\max_{j \notin S(x_{i})}{f_j\left(x_i\right)}$ for $x_i$;
						
						Correct the noisy sample $x_i$ by moving the predicted label into its candidate set $S(x_{i})$;
					}
				}
			}
		}
	\end{algorithm*}

	\subsection{Theoretical Analysis}
	In this section, we conduct a theoretical analysis to demonstrate the feasibility of IRNet. Our proof is inspired by previous works \cite{zhang2021learning, xu2022progressive} but extended to a different problem, noisy PLL. For convenience, we first introduce some necessary notations. The goal of supervised learning and noisy PLL is to learn a classifier that can make correct predictions on unseen data. Usually, we take the highest estimated probability as the predicted label $h(x)=\arg\max_{j \in \mathcal{Y}}{f_j\left(x\right)}$. We assume that all possible $h(x)$ form the hypothesis space $\mathcal{H}$.
	
	In supervised learning, each sample has a ground-truth label, i.e., $\tilde{\mathcal{D}}=\{ (x_{i}, y_{i})\}_{i=1}^{N}$. The Bayes optimal classifier $h^*(x)$ is the one that minimizes the following risk:
	\begin{equation}
	h^{*}(x)=\underset{h\in \mathcal{H}}{\arg\min}{\mathbb{E}_{(x,y)\sim \tilde{\mathcal{D}}}\mathbb{I}} (h(x) \neq y).
	\end{equation}
	
	Unlike supervised learning, each sample is associated with a set of candidate labels in noisy PLL, i.e., $\mathcal{D}_{S}=\{ \left(x_{i}, S(x_{i})\right)\}_{i=1}^{N}$. The optimal classifier $f(x)$ is the one that can minimize the risk under a suitable loss function $\mathcal{L}$:
	\begin{equation}
	f(x) = \underset{h\in \mathcal{H}}{\arg\min}{\mathbb{E}_{(x,S(x))\sim \mathcal{D}_S}\mathcal{L}} (h(x),S(x)).
	\end{equation}
	
	We assume that $\mathcal{H}$ is sufficiently complex so that $f(x)$ can approximate the Bayes optimal classifier $h^*(x)$. Let $y^x$ (or $o^x$) be the label with the highest (or second highest) posterior possibility, i.e., $y^x=\arg\max_{j \in \mathcal{Y}} \ {p(y=j|x)}$, $o^x=\arg\max_{j \in \mathcal{Y},j\neq y^x} \ p(y=j|x)$. We can measure the confidence of $x$ by the margin $u(x)=p(y^x|x)-p(o^x|x)$.
	\begin{definition}[Pure $(m,f,S)$-level set]
		\label{definition1}
		A set $L(m)= \{x| u(x) \geq m\}$ is pure for $(f,S)$ if all $x \in L(m)$ satisfy: (1) $y^x \in S(x)$; (2) $y^x=argmax_{j \in \mathcal{Y}}f_j(x)$.
	\end{definition}

	\begin{assumption}[Level set $\left(\alpha, \epsilon\right)$ consistency]
		\label{assumption1}
		Let $I(x, z)$ be an indicator function for two samples $x$ and $z$. It is equal to 1 if the more confident sample $z$ satisfies $y^z \notin S(z)$, i.e., $I(x, z) = \mathbb{I} \left[y^z \notin S(z)| u(z) \geq u(x)\right]$. Suppose there exist two constants $\alpha>0,0<\epsilon<1$ and $S_{\text{init}}(x)$ is the initial candidate set of $x$. For any mapping function $S(\cdot)$ satisfying $|S(x)|=|S_{\text{init}}(x)|$, all $x \in \mathcal{D}_{S}$ should meet the condition:
		\begin{equation}
		\left|f_j(x)-p(y=j|x)\right| < \alpha \mathbb{E}_{\left(z,S(z)\right)\sim \mathcal{D}_S} \left[ I(x, z)\right]+\frac{\epsilon}{6}.
		\end{equation}
	\end{assumption}
	Hence, the approximation error between $f_j(x)$ and $p(y=j|x)$ is controlled by the noise level of the dataset. Particularly, if $y^x$ is in $S(x)$ for all $x \in \mathcal{D}_{S}$, we will obtain a tighter constraint $\left|f_j(x)-p(y=j|x)\right| < \frac{\epsilon}{6}$.
	
	\begin{assumption}[Level set bounded distribution]
		\label{assumption2}
		Let $d(u)$ be the density function of $u(x)$. Suppose there exist two constants $0<c_*<c^*$ such that the density function $d(u)$ is bounded by $c_*<d(u)<c^*$. We denote the imbalance ratio as $l=c^*/c_*$.
	\end{assumption}
	This assumption enforces the continuity of $d(u)$. It is crucial in the analysis since it allows borrowing information from the neighborhood to help correct noisy samples. If Assumptions \ref{assumption1}$\sim$\ref{assumption2} hold, we can prove the following theorems:
	
	\begin{theorem}[One-round refinement] 
		\label{theorem1}
		Assume that there exists a boundary $\epsilon<m<1$ such that $L(m)$ is pure for $(f,S)$. Let $\epsilon<m_{\text{new}}<1$ be the new boundary. For all $x \in \mathcal{D}_{S}$ satisfying $max_{j \notin S(x)}f_j(x) - max_{j \in S(x)} f_j(x)\geq m_{\text{new}}-\epsilon$, we move $argmax_{j \notin S(x)}f_j(x) $ into $S(x)$ and move $argmin_{j \in S(x)} f_j(x)$ out of $S(x)$, thus generating the updated candidate set $S_{\text{new}}(x)$. After that, we train a new classifier $f^{\text{new}}(x)$ on the updated dataset $\mathcal{D}_{S_{\text{new}}}=\{ \left(x_{i}, S_{\text{new}}(x_{i})\right)\}_{i=1}^{N}$:
		\begin{equation}
		f^{\text{new}}(x) = \underset{h\in \mathcal{H}}{\arg\min}{\mathbb{E}_{(x,S_{\text{new}}(x))\sim \mathcal{D}_{S_{\text{new}}}}\mathcal{L}} (h(x),S_{\text{new}}(x)).
		\end{equation}
		$L(m_{\text{new}})$ is pure for $(f^{\text{new}},S_{\text{new}})$ when $m_{\text{new}}$ satisfies:
		\begin{equation}
		(1+\frac{\epsilon}{6\alpha l })(1-m) \leq 1-m_{\text{new}}\leq (1+\frac{\epsilon}{3\alpha l })(1-m).
		\end{equation}
	\end{theorem}

	\begin{theorem}[Multi-round refinement]
		\label{theorem2}
		For an initial dataset satisfying $P(y^x \notin S_{\text{init}}(x))=\eta$, we assume there exists a boundary $max(2\alpha\eta+\epsilon/3,\epsilon)<m_{\text{init}}<1$ such that $L(m_{\text{init}})$ is pure for $(f^{\text{init}}, S_{\text{init}})$. Repeat the one-round refinement in Theorem \ref{theorem1}. After $R > \frac{6l\alpha}{\epsilon}log(\frac{1-\epsilon}{1-m_{\text{init}}})$ rounds with $m_{\text{end}}=\epsilon$, we obtain the final candidate set and classifier $(S_{\text{final}},f^{\text{final}})$ that satisfy:
		\begin{equation}
		P\left(y^x \notin S_{\text{final}}(x)\right) < c^*\epsilon,
		\end{equation}
		\begin{equation}
		P\left(\underset{j \in \mathcal{Y}}{\arg\max}{f_j^{\text{final}}\left(x\right)}=h^*(x)\right) > 1 - c^*\epsilon.
		\end{equation}
	\end{theorem}
	The above theorems demonstrate that our multi-round IRNet is able to reduce the noise level of the dataset and eventually approximate the Bayes optimal classifier. The detailed proof of these theorems is shown as follows:
	
	\begin{proof_mine}[]
		\label{proof1}
		Suppose there is a boundary $\epsilon<m<1$ such that $L(m)$ is pure for $(f,S)$. Based on Definition \ref{definition1}, all $x \in L(m)$ should satisfy: (1) $y^x \in S(x)$; (2) $y^x=argmax_{j \in \mathcal{Y}}f_j(x)$. Hence, $\mathbb{E}_{(z,S(z))\sim \mathcal{D}_S} \left[\mathbb{I} \left[y^z \notin S(z)| u(z) \geq u(x)\right]\right]=0$ for all $x \in L(m)$. Combined with Assumption \ref{assumption1}, we can determine the approximation error between $f_j(x)$ and $p(y=j|x)$:
		\begin{equation}
		\label{eq16}
		\left|f_j(x)-p(y=j|x)\right|<\epsilon/6<\epsilon/2.
		\end{equation}
		Let $m_{\text{new}}$ be the new boundary satisfying $\frac{\epsilon}{6l\alpha}(1-m)<m-m_{\text{new}}<\frac{\epsilon}{3l\alpha}(1-m)$. Based on Assumption \ref{assumption2}, we can have the following result for all $x \in L(m_{\text{new}}) - L(m)$:
		\begin{align}
		\label{eq17}
		& \mathbb{E}_{(z,S(z))\sim \mathcal{D}_S} \left[\mathbb{I} \left[y^z \notin S(z)| u(z) \geq u(x)\right]\right] \nonumber\\
		= & P_z[y^z\notin S(z) | u(z)\geq u(x)] \nonumber\\
		= & \frac{P_z[y^z\notin S(z) , u(z)\geq u(x)]}{P_z[u(z)\geq u(x)]} \nonumber\\
		\leq & \frac{P_z[y^z\notin S(z) , u(z)\geq m]}{P_z[u(z)\geq u(x)]}+ \frac{P_z[y^z\notin S(z) , m\geq u(z)\geq m_{\text{new}}]}{P_z[u(z)\geq u(x)]} \nonumber\\
		= & \frac{P_z[y^z\notin S(z) , u(z)\geq m]}{P_z[u(z)\geq m]}\frac{P_z[u(z)\geq m]}{P_z[u(z)\geq u(x)]} \nonumber \\
		& + \frac{P_z[y^z\notin S(z) , m\geq u(z)\geq m_{\text{new}}]}{P_z[u(z)\geq u(x)]}\nonumber\\
		= & \mathbb{E}_{(z,S(z))\sim \mathcal{D}_S} [\mathbb{I} \left[y^z \notin S(z)| u(z)\geq m\right]]\frac{P_z[u(z)\geq m]}{P_z[u(z)\geq u(x)]} \nonumber \\
		& + \frac{P_z[y^z\notin S(z) , m\geq u(z)\geq m_{\text{new}}]}{P_z[u(z)\geq u(x)]}\nonumber\\
		= &\frac{P_z[y^z\notin S(z) , m\geq u(z)\geq m_{\text{new}}]}{P_z[u(z)\geq u(x)]}\nonumber\\
		\leq & \frac{P_z[ m\geq u(z)\geq m_{\text{new}}]}{P_z[u(z)\geq m]} \nonumber \\
		\leq & \frac{c^*(m-m_{\text{new}})}{c_*(1-m)}.
		\end{align}
		Since $m-m_{\text{new}}< \frac{\epsilon}{3l\alpha}(1-m)$, we can further relax Eq. \ref{eq17} to:
		\begin{equation}
		\label{eq18}
		\begin{split}
		& \mathbb{E}_{(z,S(z))\sim \mathcal{D}_S} \left[\mathbb{I} \left[y^z \notin S(z)| u(z) \geq u(x)\right]\right] \\
		\leq & \frac{l(m-m_{\text{new}})}{1-m} \\
		\leq & \frac{c^*}{c_*(1-m)} \frac{\epsilon}{3l\alpha}(1-m) \\
		= & \frac{\epsilon}{3\alpha}.
		\end{split}
		\end{equation}
		Then based on Assumption \ref{assumption1}, for all $x \in L(m_{\text{new}}) - L(m)$, the approximation error between $f_j(x)$ and $p(y=j|x)$ should satisfy:
		\begin{equation}
		\label{eq19}
		\begin{split}
		& |f_j(x)-p(y=j|x)| \\
		< & \alpha \mathbb{E}_{(z,S(z))\sim \mathcal{D}_S} \left[\mathbb{I} \left[y^z \notin S(z)| u(z) \geq u(x)\right]\right]+\epsilon/6 \\
		< & \alpha \frac{\epsilon}{3\alpha} +\frac{\epsilon}{6} = \frac{\epsilon}{2}.
		\end{split}
		\end{equation}
		Combining Eq. \ref{eq16} and Eq. \ref{eq19}, for all $x \in L(m_{\text{new}})$, we can prove that $|f_j(x)-p(y=j|x)| < \epsilon/2$. Then, we get:
		\begin{equation}
		\label{eq21}
		\begin{split}
		& f_{y^x}(x) - f_{j\neq y^x}(x) \\
		\geq & \left(p(y=y^x|x)-\frac{\epsilon}{2}\right) - \left(p(y=j|x)+\frac{\epsilon}{2}\right) \\
		\geq & p(y=y^x|x) - p(y=j|x)-\epsilon \\
		\geq & p(y=y^x|x) - p(y=o^x|x) - \epsilon \\
		\geq & m_{\text{new}} - \epsilon \geq 0.
		\end{split}
		\end{equation}
		It means that for all $x \in L(m_{\text{new}})$, we have $argmax_j f_j(x) = y^x$. If $argmax_j f_j(x)$ is not in $S(x)$ and $max_{j \notin S(x)}f_j(x) - max_{j \in S(x)} f_j(x)\geq m_{\text{new}}-\epsilon$, we can move $argmax_{j \notin S(x)}f_j(x) $ into $S(x)$ and move $argmin_{j \in S(x)} f_j(x)$ out of $S(x)$, thus generating the updated candidate set $S_{\text{new}}(x)$. After that, all $x \in L(m_{\text{new}})$ satisfy $|S_{\text{new}}(x)|=|S(x)|$ and $y^x \in S_{\text{new}}(x)$. Hence, we can prove that $L(m_{\text{new}})$ is pure for $(f,S_{\text{new}})$. 
		
		Then, we train a new classifier on the updated dataset $\mathcal{D}_{S_{\text{new}}}$, i.e., $f^{\text{new}}(x) = argmin_{h\in \mathcal{H}}{\mathbb{E}_{(x,S_{\text{new}}(x))\sim \mathcal{D}_{S_{\text{new}}}}\mathcal{L}} (h(x),S_{\text{new}}(x))$. Repeating the proof of Theorem \ref{theorem1}, we can prove that all $x \in L(m_{\text{new}})$ should satisfy $|f_j^{\text{new}}(x)-p(y=j|x)|<\epsilon/2$. Based on Eq. \ref{eq21}, we have $argmax_j f^{\text{new}}_j(x) = y^x$. Since $y^x$ is already in $S^{\text{new}}(x)$, we can prove that $L(m_{\text{new}})$ is pure for $(f^{\text{new}},S_{\text{new}})$.
	\end{proof_mine}

	\begin{proof_mine}[]
		\label{proof2}
		For an initial dataset satisfying $P(y^x \notin S_{\text{init}}(x))=\eta$, all $x \in \mathcal{D}_{S_{\text{init}}}$ should satisfy:
		\begin{equation}
		\mathbb{E}_{(z,S_{\text{init}}(z))\sim \mathcal{D}_{S_{\text{init}}}} \left[\mathbb{I}\left[y^z \notin S_{\text{init}}(z)| u(z) \geq u(x)\right]\right]=\eta.
		\end{equation}
		Based on Assumption \ref{assumption1}, we can determine that the approximation error between $f_j^{\text{init}}(x)$ and $p(y=j|x)$:
		\begin{equation}
		|f_j^{\text{init}}(x)-p(y=j|x)|<\alpha \eta+\frac{\epsilon}{6}.
		\end{equation}
		Let $m_{\text{init}}$ be the initial boundary. According to Eq. \ref{eq21}, $L(m_{\text{init}})$ is pure only if $|f_j^{\text{init}}(x)-p(y=j|x)|<\frac{m_{\text{init}}}{2}$. Therefore, we choose $m_{\text{init}}$ from $max(2\alpha\eta +\frac{\epsilon}{3},\epsilon)<m_{\text{init}}<1$. 
		
		According to Theorem \ref{theorem1}, the updated boundary $m_{\text{new}}$ should satisfy $\left(1+\frac{\epsilon}{6l\alpha}\right)(1-m) \leq 1-m_{\text{new}}$. Therefore, we need $R$ rounds of refinement to update the boundary from $m_{\text{init}}$ to $m_{\text{end}}=\epsilon$. The value of $R$ is calculated as follows:
		\begin{equation}
		\begin{split}
		\left(1+\frac{\epsilon}{6l\alpha}\right)^R\left(1-m_{\text{init}}\right)&>1-\epsilon \\
		\left(1+\frac{\epsilon}{6l\alpha}\right)^R&>\frac{1-\epsilon}{1-m_{\text{init}}} \\
		R log\left(1+\frac{\epsilon}{6l\alpha}\right)&>log\left(\frac{1-\epsilon}{1-m_{\text{init}}}\right).
		\end{split}
		\end{equation}
		Then, we have:
		\begin{equation}
		\begin{split}
		R & >log\left(\frac{1-\epsilon}{1-m_{\text{init}}}\right) / log\left(1+\frac{\epsilon}{6l\alpha}\right) \\
		&> log\left(\frac{1-\epsilon}{1-m_{\text{init}}}\right) / \frac{\epsilon}{6l\alpha}. 
		\end{split}
		\end{equation}
		After multiple rounds of refinement, we reach the final boundary $m_{\text{end}}=\epsilon$. The final candidate set $S_{\text{final}}$ and the final classifier $f^{\text{final}}$ should meet the following conditions:
		\begin{equation}
		\begin{split}
		P\left(y^x \notin S_{\text{final}}(x)\right) & \leq P(u(x) <m_{\text{end}}) \\
		&= P(u(x) < \epsilon) \\
		&< c^* \epsilon,
		\end{split}
		\end{equation}
		\begin{equation}
		\begin{split}
		P\left(\underset{j \in \mathcal{Y}}{\arg\max}{f_j^{\text{final}}\left(x\right)} \neq h^*(x)\right) & \leq P(u(x) <m_{\text{end}}) \\
		&= P(u(x) < \epsilon) \\
		&< c^* \epsilon.
		\end{split}
		\end{equation}
		Therefore, we can prove that:
		\begin{equation}
		\begin{split}
		& P\left(\underset{j \in \mathcal{Y}}{\arg\max}{f_j^{\text{final}}\left(x\right)} = h^*(x)\right) \\
		&= 1-P\left(\underset{j \in \mathcal{Y}}{\arg\max}{f_j^{\text{final}}\left(x\right)} \neq h^*(x)\right) \\
		&> 1 - c^* \epsilon.
		\end{split}
		\end{equation}
	\end{proof_mine}
	
	It should be noticed that theory is slightly different from practice. In our implementation, we correct the noisy sample by moving the predicted label into the candidate set. But to reduce the difficulty of the theoretical proof, we further remove the candidate label with the lowest confidence (i.e., $\arg\min_{j \in S(x)}{f_j\left(x\right)}$) to keep the number of candidate labels consistent. We compare these strategies in Section \ref{sec:role_of_swapping}. Experimental results show that this removal operation leads to a slight decrease in classification performance. Therefore, we do not exploit the removal operation in practice.
	
	\section{Experimental Databases and Setup}
	\label{sec4}

	\subsection{Corpus Description}
	\label{Sec4-1}
	We conduct experiments on three benchmark datasets, including CIFAR-10 \cite{krizhevsky2009learning}, CIFAR-100 \cite{krizhevsky2009learning}, and Kuzushiji-MNIST \cite{clanuwat2018deep}. We manually corrupt these datasets into noisy partially labeled versions. To form the candidate set for clean samples, we flip incorrect labels to false positive labels with a probability $q$ and aggregate the flipped ones with the ground-truth label, in line with previous works \cite{lv2020progressive, wen2021leveraged}. After that, each sample has a probability $\eta$ of being a noisy sample. To form the candidate set for each noisy sample, we further select a negative label from the non-candidate set, move it into the candidate set, and move the ground-truth label out of the candidate set. In this paper, we denote the probability $q$ as the ambiguity level and the probability $\eta$ as the noise level. The ambiguity level controls the number of candidate labels, and the noise level controls the percentage of noisy samples. Since CIFAR-100 has 10 times more labels than other datasets, we consider $q\in \left\{0.01, 0.03, 0.05\right\}$ for CIFAR-100 and $q\in \left\{0.1, 0.3, 0.5\right\}$ for others.
	
	In addition to the aforementioned datasets, we also provide practical application scenarios for noisy PLL. Facial expression recognition is important for enhancing user experience in human-computer interaction. Due to the complexity of emotions, different annotators assign distinct labels to the same image, resulting in a possible lack of ground-truth labels in the candidate set and causing the noisy PLL problem. RAF-DB \cite{li2017reliable} is one of the most commonly used datasets in this field, where each sample is labeled by approximately 40 annotators. This paper leverages the proportion of dominant emotions as the reliability score and selects samples with high scores to form the test set. For example, if a sample is labeled \emph{happy} 10 times, \emph{sad} 8 times and \emph{neutral} 2 times, its reliability score is $10/(10+8+2)=0.5$. This process ensures that different methods are evaluated on reliable samples and improves the persuasiveness of evaluation results.
	
	\begin{table*}[htbp]
		\centering
		\renewcommand\tabcolsep{5pt}
		\renewcommand\arraystretch{1.20}
		\caption{Compatibility of IRNet with existing PLL methods. In this table, we report the inductive performance (mean$\pm$std).}
		\label{Table2}
		\begin{tabular}{lc|ccc|ccc|ccc}
			\hline
			\multirow{3}{*}{PLL}& \multirow{3}{*}{IRNet}& \multicolumn{9}{c}{CIFAR-10} \\
			\cline{3-11}
			& & \multicolumn{3}{c|}{$q=0.1$} & \multicolumn{3}{c|}{$q=0.3$} & \multicolumn{3}{c}{$q=0.5$} \\ \cline{3-11}
			& & $\eta=0.1$ & $\eta=0.2$ & $\eta=0.3$ & $\eta=0.1$ & $\eta=0.2$ & $\eta=0.3$ & $\eta=0.1$ & $\eta=0.2$ & $\eta=0.3$ \\
			\hline
			CC & $\times$& 89.16$\pm$0.23& 85.72$\pm$0.27& 84.46$\pm$0.14& 86.14$\pm$0.22& 81.48$\pm$0.00& 77.04$\pm$0.00& 80.76$\pm$0.52& 73.56$\pm$0.00& 62.95$\pm$0.59\\
			CC & $\surd$& \cellcolor{black!18}90.91$\pm$0.05& \cellcolor{black!18}90.56$\pm$0.03& \cellcolor{black!18}89.92$\pm$0.00& \cellcolor{black!18}90.45$\pm$0.04& \cellcolor{black!18}89.45$\pm$0.19& \cellcolor{black!18}87.66$\pm$0.24& \cellcolor{black!18}87.82$\pm$0.10& \cellcolor{black!18}84.97$\pm$0.35& \cellcolor{black!18}76.16$\pm$0.49\\
			
			RC & $\times$& 89.80$\pm$0.19& 87.11$\pm$0.02& 85.81$\pm$0.37& 88.84$\pm$0.01& 85.96$\pm$0.09& 82.23$\pm$0.48& 86.98$\pm$0.00& 81.75$\pm$0.18& 73.63$\pm$0.20\\
			RC & $\surd$& \cellcolor{black!18}91.06$\pm$0.13& \cellcolor{black!18}90.03$\pm$0.17& \cellcolor{black!18}89.92$\pm$0.28& \cellcolor{black!18}90.51$\pm$0.39& \cellcolor{black!18}89.85$\pm$0.36& \cellcolor{black!18}89.36$\pm$0.24& \cellcolor{black!18}89.63$\pm$0.23& \cellcolor{black!18}87.54$\pm$0.23& \cellcolor{black!18}81.08$\pm$0.06\\
			
			LOG & $\times$& 88.83$\pm$0.17& 85.65$\pm$0.08& 83.99$\pm$0.02& 86.39$\pm$0.24& 81.01$\pm$0.11& 77.53$\pm$0.19& 81.59$\pm$0.07& 73.12$\pm$0.20& 63.89$\pm$0.61\\
			LOG & $\surd$& \cellcolor{black!18}90.84$\pm$0.15& \cellcolor{black!18}90.53$\pm$0.16& \cellcolor{black!18}89.52$\pm$0.22& \cellcolor{black!18}89.67$\pm$0.16& \cellcolor{black!18}89.70$\pm$0.28& \cellcolor{black!18}87.78$\pm$0.00& \cellcolor{black!18}88.37$\pm$0.25& \cellcolor{black!18}85.12$\pm$0.24& \cellcolor{black!18}77.47$\pm$0.39\\
			
			PRODEN & $\times$& 89.78$\pm$0.20& 87.33$\pm$0.04& 84.33$\pm$0.08& 88.86$\pm$0.05& 85.51$\pm$0.12& 82.02$\pm$0.00& 87.15$\pm$0.22& 81.95$\pm$0.42& 73.33$\pm$0.38\\
			PRODEN & $\surd$& \cellcolor{black!18}90.86$\pm$0.05& \cellcolor{black!18}89.70$\pm$0.17& \cellcolor{black!18}89.94$\pm$0.05& \cellcolor{black!18}90.59$\pm$0.16& \cellcolor{black!18}89.98$\pm$0.10& \cellcolor{black!18}89.06$\pm$0.03& \cellcolor{black!18}89.55$\pm$0.04& \cellcolor{black!18}87.53$\pm$0.10& \cellcolor{black!18}78.48$\pm$0.67\\
			
			LWC & $\times$& 89.05$\pm$0.10& 85.90$\pm$0.08& 83.62$\pm$0.10& 88.22$\pm$0.03& 84.42$\pm$0.11& 80.97$\pm$0.16& 86.19$\pm$0.10& 79.54$\pm$0.10& 69.95$\pm$0.52\\
			LWC & $\surd$& \cellcolor{black!18}90.38$\pm$0.03& \cellcolor{black!18}89.27$\pm$0.21& \cellcolor{black!18}88.98$\pm$0.33& \cellcolor{black!18}90.13$\pm$0.03& \cellcolor{black!18}89.36$\pm$0.08& \cellcolor{black!18}88.83$\pm$0.18& \cellcolor{black!18}88.59$\pm$0.03& \cellcolor{black!18}87.36$\pm$0.05& \cellcolor{black!18}77.32$\pm$1.43\\
			
			PiCO &$\times$ & 90.78$\pm$0.24 & 87.27$\pm$0.11 & 84.96$\pm$0.12 & 89.71$\pm$0.18 & 85.78$\pm$0.23 & 82.25$\pm$0.32 & 88.11$\pm$0.29 & 82.41$\pm$0.30 & 68.75$\pm$2.62\\
			PiCO &$\surd$ & \cellcolor{black!18}92.69$\pm$0.16 & \cellcolor{black!18}92.12$\pm$0.09 & \cellcolor{black!18}92.38$\pm$0.21 & \cellcolor{black!18}92.10$\pm$0.02 & \cellcolor{black!18}91.69$\pm$0.26 & \cellcolor{black!18}91.35$\pm$0.08 & \cellcolor{black!18}91.51$\pm$0.05 & \cellcolor{black!18}90.76$\pm$0.10 & \cellcolor{black!18}86.19$\pm$0.41\\ 
			
			\hline
			\hline
			\multirow{3}{*}{PLL}& \multirow{3}{*}{IRNet}& \multicolumn{9}{c}{CIFAR-100} \\
			\cline{3-11}
			& & \multicolumn{3}{c|}{$q=0.01$} & \multicolumn{3}{c|}{$q=0.03$} & \multicolumn{3}{c}{$q=0.05$} \\ \cline{3-11}
			& & $\eta=0.1$ & $\eta=0.2$ & $\eta=0.3$ & $\eta=0.1$ & $\eta=0.2$ & $\eta=0.3$ & $\eta=0.1$ & $\eta=0.2$ & $\eta=0.3$ \\
			\hline
			CC & $\times$& 67.83$\pm$0.33& 61.70$\pm$0.00& 58.80$\pm$0.55& 66.92$\pm$0.12& 60.33$\pm$0.00& 57.36$\pm$0.02& 65.80$\pm$0.26& 60.06$\pm$0.03& 55.64$\pm$0.41\\
			CC & $\surd$& \cellcolor{black!18}71.85$\pm$0.24& \cellcolor{black!18}69.54$\pm$0.03& \cellcolor{black!18}67.40$\pm$0.29& \cellcolor{black!18}70.65$\pm$0.11& \cellcolor{black!18}69.10$\pm$0.05& \cellcolor{black!18}65.91$\pm$0.05& \cellcolor{black!18}70.43$\pm$0.44& \cellcolor{black!18}68.16$\pm$0.34& \cellcolor{black!18}65.61$\pm$0.11\\
			
			RC & $\times$& 67.52$\pm$0.02& 62.43$\pm$0.15& 59.35$\pm$0.26& 67.27$\pm$0.16& 62.84$\pm$0.00& 60.00$\pm$0.03& 67.08$\pm$0.02& 63.12$\pm$0.08& 59.35$\pm$0.34\\
			RC & $\surd$& \cellcolor{black!18}70.45$\pm$0.18& \cellcolor{black!18}68.52$\pm$0.12& \cellcolor{black!18}66.90$\pm$0.02& \cellcolor{black!18}70.72$\pm$0.05& \cellcolor{black!18}68.16$\pm$0.25& \cellcolor{black!18}65.75$\pm$0.25& \cellcolor{black!18}70.31$\pm$0.07& \cellcolor{black!18}68.15$\pm$0.13& \cellcolor{black!18}65.19$\pm$0.10\\
			
			LOG & $\times$& 67.67$\pm$0.06& 62.34$\pm$0.00& 58.06$\pm$0.04& 66.74$\pm$0.11& 60.84$\pm$0.34& 57.05$\pm$0.03& 65.66$\pm$0.40& 60.17$\pm$0.09& 56.76$\pm$0.00\\
			LOG & $\surd$& \cellcolor{black!18}71.85$\pm$0.29& \cellcolor{black!18}69.36$\pm$0.14& \cellcolor{black!18}67.64$\pm$0.03& \cellcolor{black!18}71.32$\pm$0.17& \cellcolor{black!18}69.03$\pm$0.09& \cellcolor{black!18}66.25$\pm$0.05& \cellcolor{black!18}70.52$\pm$0.21& \cellcolor{black!18}68.28$\pm$0.00& \cellcolor{black!18}65.54$\pm$0.15\\
			
			PRODEN & $\times$& 67.87$\pm$0.38& 62.57$\pm$0.15& 59.56$\pm$0.25& 67.08$\pm$0.22& 62.56$\pm$0.05& 59.47$\pm$0.24& 66.83$\pm$0.18& 63.05$\pm$0.11& 59.46$\pm$0.13\\
			PRODEN & $\surd$& \cellcolor{black!18}70.72$\pm$0.17& \cellcolor{black!18}68.31$\pm$0.21& \cellcolor{black!18}66.26$\pm$0.10& \cellcolor{black!18}70.66$\pm$0.11& \cellcolor{black!18}68.12$\pm$0.22& \cellcolor{black!18}65.51$\pm$0.05& \cellcolor{black!18}70.43$\pm$0.18& \cellcolor{black!18}68.11$\pm$0.09& \cellcolor{black!18}65.07$\pm$0.05\\
			
			LWC & $\times$& 67.47$\pm$0.33& 61.78$\pm$0.12& 58.53$\pm$0.16& 66.84$\pm$0.00& 62.21$\pm$0.16& 58.55$\pm$0.09& 66.19$\pm$0.36& 62.02$\pm$0.02& 58.74$\pm$0.06\\
			LWC & $\surd$& \cellcolor{black!18}71.10$\pm$0.17& \cellcolor{black!18}68.69$\pm$0.02& \cellcolor{black!18}66.25$\pm$0.02& \cellcolor{black!18}70.49$\pm$0.10& \cellcolor{black!18}68.53$\pm$0.07& \cellcolor{black!18}66.08$\pm$0.09& \cellcolor{black!18}70.60$\pm$0.03& \cellcolor{black!18}68.34$\pm$0.18& \cellcolor{black!18}65.47$\pm$0.18\\
			
			PiCO &$\times$& 68.27$\pm$0.08 & 62.24$\pm$0.31 & 58.97$\pm$0.09 & 67.38$\pm$0.09 & 62.01$\pm$0.33 & 58.64$\pm$0.28 & 67.52$\pm$0.43 & 61.52$\pm$0.28 & 58.18$\pm$0.65\\
			PiCO &$\surd$& \cellcolor{black!18}71.17$\pm$0.14 & \cellcolor{black!18}70.10$\pm$0.28 & \cellcolor{black!18}68.77$\pm$0.28 & \cellcolor{black!18}71.01$\pm$0.43 & \cellcolor{black!18}70.15$\pm$0.17 & \cellcolor{black!18}68.18$\pm$0.30 & \cellcolor{black!18}70.73$\pm$0.09 & \cellcolor{black!18}69.33$\pm$0.51 & \cellcolor{black!18}68.09$\pm$0.12\\ 
			
			\hline
			\hline
			\multirow{3}{*}{PLL}& \multirow{3}{*}{IRNet}& \multicolumn{9}{c}{Kuzushiji-MNIST} \\
			\cline{3-11}
			& & \multicolumn{3}{c|}{$q=0.1$} & \multicolumn{3}{c|}{$q=0.3$} & \multicolumn{3}{c}{$q=0.5$} \\ \cline{3-11}
			& & $\eta=0.1$ & $\eta=0.2$ & $\eta=0.3$ & $\eta=0.1$ & $\eta=0.2$ & $\eta=0.3$ & $\eta=0.1$ & $\eta=0.2$ & $\eta=0.3$ \\
			\hline
			CC & $\times$& 97.17$\pm$0.00& 96.90$\pm$0.15& 95.44$\pm$0.17& 96.69$\pm$0.04& 96.02$\pm$0.12& 93.34$\pm$0.21& 94.92$\pm$0.24& 91.25$\pm$0.31& 84.92$\pm$0.08\\
			CC & $\surd$& \cellcolor{black!18}98.29$\pm$0.04& \cellcolor{black!18}98.13$\pm$0.02& \cellcolor{black!18}98.13$\pm$0.12& \cellcolor{black!18}98.02$\pm$0.00& \cellcolor{black!18}97.53$\pm$0.07& \cellcolor{black!18}97.10$\pm$0.21& \cellcolor{black!18}97.29$\pm$0.33& \cellcolor{black!18}95.76$\pm$0.27& \cellcolor{black!18}93.97$\pm$0.08\\
			
			RC & $\times$& 97.61$\pm$0.00& 97.10$\pm$0.08& 96.28$\pm$0.05& 97.52$\pm$0.05& 96.72$\pm$0.03& 94.98$\pm$0.06& 97.02$\pm$0.07& 94.12$\pm$0.16& 90.62$\pm$0.13\\
			RC & $\surd$& \cellcolor{black!18}98.07$\pm$0.07& \cellcolor{black!18}97.88$\pm$0.08& \cellcolor{black!18}97.87$\pm$0.04& \cellcolor{black!18}97.95$\pm$0.00& \cellcolor{black!18}97.88$\pm$0.00& \cellcolor{black!18}97.33$\pm$0.03& \cellcolor{black!18}97.68$\pm$0.18& \cellcolor{black!18}96.70$\pm$0.13& \cellcolor{black!18}94.59$\pm$0.12\\
			
			LOG & $\times$& 97.56$\pm$0.10& 97.16$\pm$0.01& 96.25$\pm$0.30& 97.08$\pm$0.28& 94.91$\pm$0.11& 93.73$\pm$0.46& 94.56$\pm$0.20& 91.09$\pm$0.04& 84.44$\pm$0.30\\
			LOG & $\surd$& \cellcolor{black!18}98.08$\pm$0.04& \cellcolor{black!18}97.95$\pm$0.11& \cellcolor{black!18}97.85$\pm$0.06& \cellcolor{black!18}98.06$\pm$0.05& \cellcolor{black!18}97.69$\pm$0.07& \cellcolor{black!18}96.75$\pm$0.08& \cellcolor{black!18}97.39$\pm$0.32& \cellcolor{black!18}95.88$\pm$0.58& \cellcolor{black!18}93.34$\pm$0.79\\
			
			PRODEN & $\times$& 97.58$\pm$0.00& 97.21$\pm$0.03& 96.39$\pm$0.03& 97.38$\pm$0.10& 96.47$\pm$0.33& 94.97$\pm$0.11& 96.69$\pm$0.05& 94.60$\pm$0.01& 90.91$\pm$0.26\\
			PRODEN & $\surd$& \cellcolor{black!18}98.07$\pm$0.05& \cellcolor{black!18}98.11$\pm$0.04& \cellcolor{black!18}97.86$\pm$0.10& \cellcolor{black!18}98.07$\pm$0.00& \cellcolor{black!18}98.03$\pm$0.23& \cellcolor{black!18}97.39$\pm$0.05& \cellcolor{black!18}97.57$\pm$0.34& \cellcolor{black!18}96.71$\pm$0.34& \cellcolor{black!18}94.27$\pm$0.11\\
			
			LWC & $\times$& 97.25$\pm$0.04& 96.60$\pm$0.01& 95.66$\pm$0.15& 97.11$\pm$0.10& 96.44$\pm$0.28& 93.95$\pm$0.03& 96.40$\pm$0.05& 93.42$\pm$0.16& 88.53$\pm$0.39\\
			LWC & $\surd$& \cellcolor{black!18}97.95$\pm$0.11& \cellcolor{black!18}97.89$\pm$0.11& \cellcolor{black!18}97.91$\pm$0.11& \cellcolor{black!18}98.05$\pm$0.04& \cellcolor{black!18}97.74$\pm$0.05& \cellcolor{black!18}97.08$\pm$0.53& \cellcolor{black!18}97.31$\pm$0.15& \cellcolor{black!18}96.91$\pm$0.48& \cellcolor{black!18}94.55$\pm$0.24\\
			
			PiCO &$\times$& 97.55$\pm$0.04 & 96.95$\pm$0.16 & 96.53$\pm$0.18 & 97.34$\pm$0.05 & 96.57$\pm$0.06 & 95.34$\pm$0.34 & 96.65$\pm$0.09 & 95.17$\pm$0.34 & 90.87$\pm$1.75\\
			PiCO &$\surd$ & \cellcolor{black!18}98.07$\pm$0.07 & \cellcolor{black!18}97.96$\pm$0.03 & \cellcolor{black!18}97.95$\pm$0.07 & \cellcolor{black!18}97.97$\pm$0.03 & \cellcolor{black!18}97.96$\pm$0.09 & \cellcolor{black!18}97.87$\pm$0.10 & \cellcolor{black!18}97.86$\pm$0.23 & \cellcolor{black!18}97.34$\pm$0.22 & \cellcolor{black!18}95.84$\pm$0.34\\
			
			\hline
		\end{tabular}
	\end{table*}
	
	\subsection{Baselines}
	This paper considers the following state-of-the-art methods as baselines. For a fair comparison, we reproduce all baselines using the same candidate label generation strategy, backbone, and data augmentation approach.
	
	\textbf{LOG} \cite{feng2020learning} is a strong benchmark model that exploits an upper-bound loss to improve classification performance.
	
	\textbf{CC} \cite{feng2020provably} is a classifier-consistent method that leverages the transition matrix to form an empirical risk estimator.
	
	\textbf{RC} \cite{feng2020provably} is a risk-consistent method that uses importance reweighting to approximate the optimal classifier.
	
	\textbf{PRODEN} \cite{lv2020progressive} is a self-training technique that progressively identifies the ground-truth label using model outputs.
	
	\textbf{LWC} \cite{wen2021leveraged} considers the trade-off between losses on candidate labels and non-candidate labels.
	
	
	
	\textbf{PiCO} \cite{wang2022pico} is a contrastive-learning-based PLL method that exploits a prototype-based disambiguation algorithm to identify the ground-truth label from the candidate set.
	
	\textbf{PiCO+} \cite{wang2023pico+} extends PiCO to deal with noisy PLL. Inspired by DivideMix \cite{li2020dividemix}, it performs distance-based clean sample selection and learns robust classifiers in a semi-supervised learning manner. 
	
	\textbf{PLCR} \cite{wu2022revisiting} performs consistency regularization on the candidate set and employs supervised learning on the non-candidate set. Similar to smoothness constraints in IRNet, this consistency regularization can also improve the reliability of prediction results under noisy conditions.
	
	Lv et al. \cite{lv2023robustness} rely on noise-robust loss functions to solve noisy PLL. To introduce them clearly, we convert the ground-truth label $y$ into its one-hot version $y_j(x), j \in [1, C]$, where $y_j(x)$ is equal to 1 if $j=y$ and 0 otherwise.
	
	\textbf{MAE} \cite{ghosh2017robust} is bounded and symmetric. Previous works have proved that this loss function is robust to label noise:
	\begin{equation}
	\mathcal{L}_{\text{MAE}}=-\sum\limits_{j=1}^{C}\lVert y_j(x) - f_j(x) \rVert_1.
	\end{equation}
	
	\textbf{MSE} \cite{ghosh2017robust} is bounded but not symmetric. Same with MAE, it is robust to label noise:
	\begin{equation}
	\mathcal{L}_{\text{MSE}}=-\sum\limits_{j=1}^{C}\lVert y_j(x) - f_j(x) \rVert_2^2.
	\end{equation}
	
	\textbf{SCE} \cite{wang2019symmetric} combines RCE with CCE via $\lambda_c$ and $\lambda_r$. This loss function exploits the benefits of the noise-robustness provided by RCE and the implicit weighting scheme of CCE:
	\begin{equation}
	\mathcal{L}_{\text{CCE}}=-\sum\limits_{j=1}^{C}y_j(x)\log f_j(x),
	\end{equation}
	\begin{equation}
	\mathcal{L}_{\text{RCE}}=-\sum\limits_{j=1}^{C}f_j(x)\log y_j(x),
	\end{equation}
	\begin{equation}
	\mathcal{L}_{\text{SCE}}=\lambda_c \mathcal{L}_{\text{CCE}} + \lambda_r \mathcal{L}_{\text{RCE}}.
	\end{equation}
	
	\textbf{GCE \cite{zhang2018generalized}} adopts the negative Box-Cox transformation and utilizes a hyper-parameter $\lambda_g$ to balance MAE and CCE:
	\begin{equation}
	\mathcal{L}_{\text{GCE}}=\sum\limits_{j=1}^{C}y_j(x)\left(\frac{1-f_j(x)^{\lambda_g}}{\lambda_g}\right).
	\end{equation}

	\subsection{Implementation Details}
	\label{sec:implementation_details}
	IRNet mainly contains two user-specific parameters: the margin $\tau_\epsilon$ and the number of augmentations $K$. We select $\tau_\epsilon$ from $\{0.0001, 0.001, 0.002, 0.004, 0.008, 0.016\}$ and $K$ from $\{1, 2\}$. To optimize all trainable parameters, we exploit the SGD optimizer and set the maximum number of epochs to $E_{\max}=1000$. We set the initial learning rate to $0.01$ and adjust it using the cosine scheduler. Train/test splits are provided in our experimental datasets. We conduct each experiment three times with distinct random seeds and report the average results on the test set. All experiments are implemented with PyTorch \cite{paszke2019pytorch} and carried out with NVIDIA Tesla V100 GPUs.
	
	\section{Results and Discussion}
	\label{sec5}
	In this section, we first report inductive and transductive results under varying ambiguity and noise levels. Then, we show the noise robustness of our method and reveal the importance of smoothness constraints. Next, we conduct parameter sensitivity analysis and study the role of swapping in label correction. Following that, we visualize the latent representations for qualitative analysis. Finally, we present practical application scenarios of noisy PLL.

	\begin{table}[t]
		\centering
		\renewcommand\arraystretch{1.20}
		\caption{Inductive performance of different noise-robust techniques. The noise level of these datasets is fixed to $\eta=0.3$. In this table, the best performance is highlighted in bold.}
		\label{Table3}
		\begin{tabular}{c|c|c|c|c}
			\hline
			Dataset & Method & $q=0.1$ & $q=0.3$ & $q=0.5$ \\
			\hline
			
			\multirow{6}{*}{CIFAR-10} &MAE &89.82 &86.71 &76.42 \\ 
			&MSE 	&86.16 &82.44 &71.25 \\ 
			&SCE 	&89.04 &84.42 &59.45 \\ 
			&GCE 	&88.77 &85.98 &77.95 \\ 
			&PLCR   &89.44 &85.81 &75.47 \\
			&PiCO+ 	&{92.19} &{89.86} &{83.99}\\
			&\textbf{IRNet} &\textbf{92.38} &\textbf{91.35} &\textbf{86.19} \\ 
			\hline
			
			\multirow{6}{*}{{\begin{tabular}[c]{@{}c@{}}Kuzushiji- \\ MNIST\end{tabular} }} &MAE &97.38 &96.52 &93.20 \\ 
			&MSE &96.59 &95.03 &86.87 \\ 
			&SCE &97.30 &96.09 &89.30 \\ 
			&GCE &97.51 &96.16 &92.55 \\ 
			&PLCR &97.83  &97.14  &92.98 \\
			&PiCO+ &95.48 & 94.65 &83.52\\
			&\textbf{IRNet} &\textbf{97.95} &\textbf{97.87} &\textbf{95.84} \\ 
			\hline
			\hline
			
			Dataset & Method & $q=0.01$ & $q=0.03$ & $q=0.05$ \\
			\hline
			
			\multirow{6}{*}{CIFAR-100} &MAE &21.97 &18.27 &19.26 \\ 
			&MSE &57.09 &51.47 &29.26 \\ 
			&SCE &40.47 &27.00 &18.63 \\ 
			&GCE &62.13 &57.95 &41.23 \\ 
			&PLCR &66.95 &66.92  &66.76 \\
			&PiCO+ &66.04 & 64.68 & 62.26 \\
			&\textbf{IRNet} &\textbf{68.77} &\textbf{68.18} &\textbf{68.09} \\ 
			\hline
			
		\end{tabular}
	\end{table}

	\subsection{Inductive Performance}
	Tables \ref{Table2} and \ref{Table3} report the inductive performance of different methods. Inductive results indicate the classification accuracy on the test set \cite{lyu2019gm, lv2020progressive}. Due to the small variance under distinct random seeds (see Table \ref{Table2}), we only report the average results in Table \ref{Table3}. 
	
	Experimental results in Table \ref{Table2} demonstrate that IRNet always brings performance improvements to existing PLL methods, verifying its effectiveness and compatibility. Meanwhile, the combination of IRNet and PiCO generally achieves better performance than other combinations. The reason lies in that PiCO is a natural co-training \cite{han2018co} framework with two branches: one for classification and one for clustering. According to previous works \cite{yu2019does}, co-training networks are more robust to noisy samples. Therefore, we combine IRNet with PiCO by default in the following experiments. Meanwhile, the advantage of our method becomes more significant with increasing noise levels, which further verifies the noise robustness of our method.
	
	In Table \ref{Table3}, IRNet achieves better performance than other noise-robust baselines. Lv et al. \cite{lv2023robustness} exploit bounded losses to avoid overemphasis on noisy samples, which also causes them to fail to use the useful information in noisy samples. Differently, IRNet can purify these samples, allowing us to leverage them to learn more discriminative classifiers. PLCR uses consistency regularization to improve the reliability of prediction results without any label correction process. Differently, IRNet progressively purifies noisy samples to clean ones through multi-round refinement. PiCO+ relies on distance-based noisy sample detection. Compared with PiCO+, IRNet detects noisy samples and corrects labels through different strategies with strong motivation and theoretical guarantees, thereby achieving better performance.

	\begin{figure*}[t]
		\begin{center}
			\subfigure[CIFAR-10 ($q=0.1$)]{
				\label{Figure4-1}
				\centering
				\includegraphics[width=0.316\linewidth] {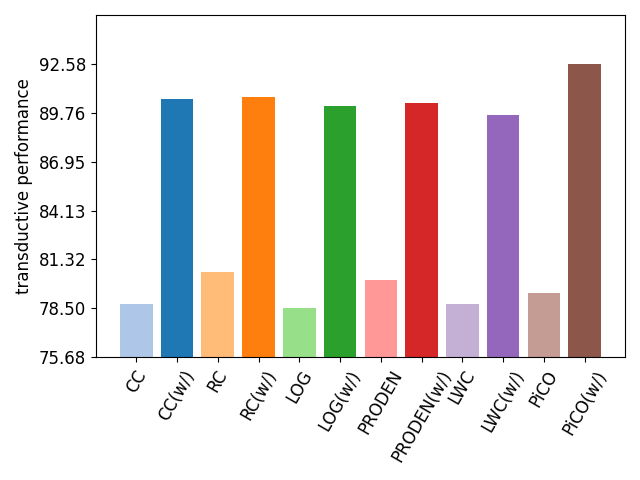}
			}
			\subfigure[CIFAR-100 ($q=0.01$)]{
				\label{Figure4-2}
				\centering
				\includegraphics[width=0.316\linewidth] {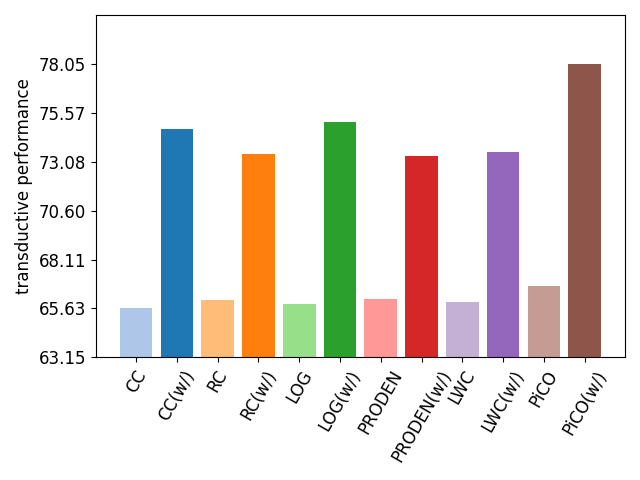}
			}
			\subfigure[Kuzushiju-MNIST ($q=0.1$)]{
				\label{Figure4-4}
				\centering
				\includegraphics[width=0.316\linewidth] {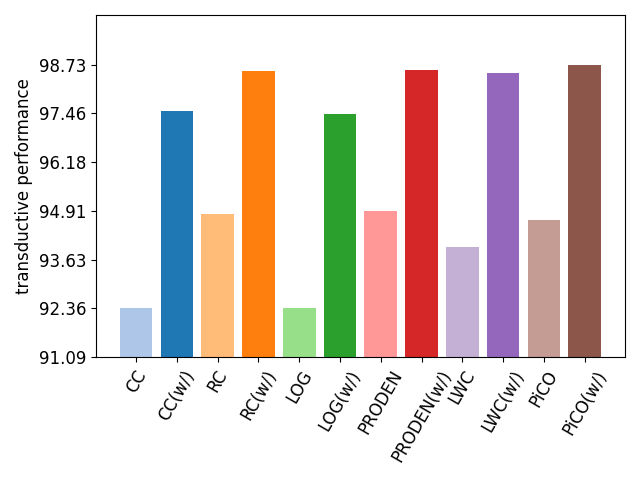}
			}
			
			\subfigure[CIFAR-10 ($q=0.3$)]{
				\label{Figure4-5}
				\centering
				\includegraphics[width=0.316\linewidth] {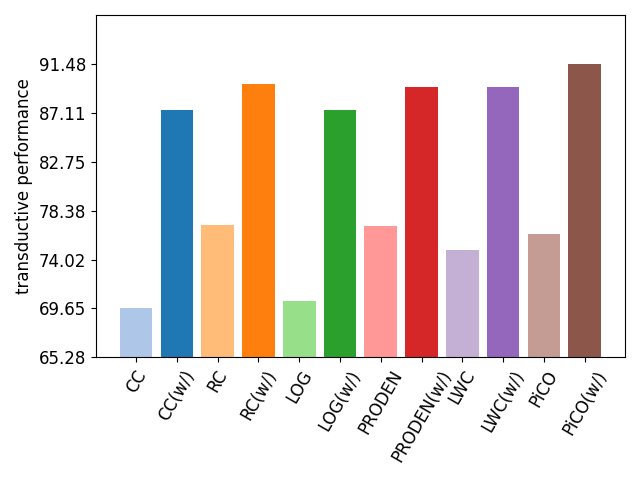}
			}
			\subfigure[CIFAR-100 ($q=0.03$)]{
				\label{Figure4-6}
				\centering
				\includegraphics[width=0.316\linewidth] {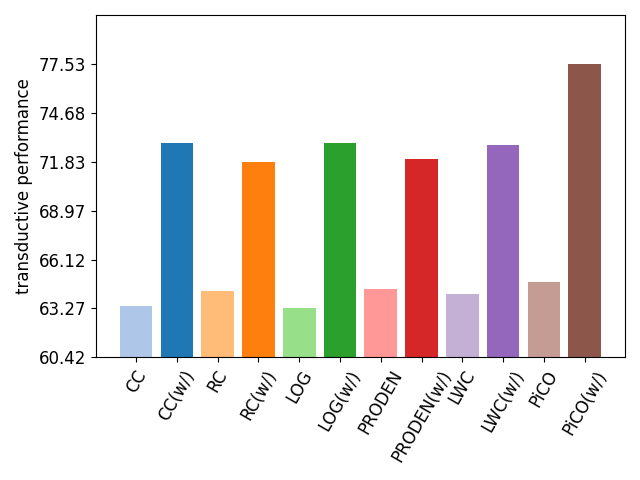}
			}
			\subfigure[Kuzushiju-MNIST ($q=0.3$)]{
				\label{Figure4-8}
				\centering
				\includegraphics[width=0.316\linewidth] {image/disambiguity_kmnist_q03n03}
			}
			
			\subfigure[CIFAR-10 ($q=0.5$)]{
				\label{Figure4-9}
				\centering
				\includegraphics[width=0.316\linewidth] {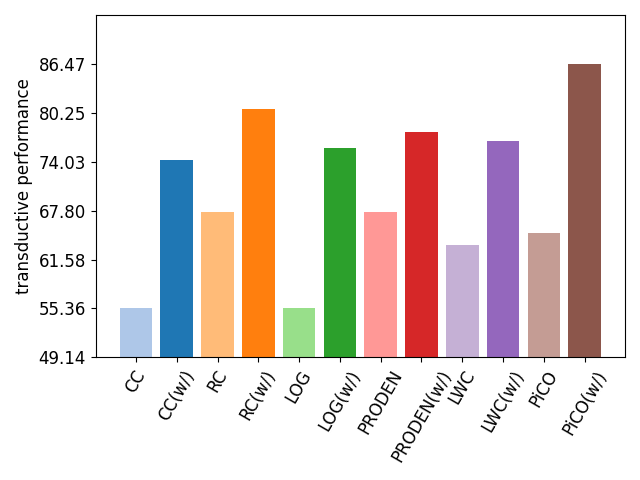}
			}
			\subfigure[CIFAR-100 ($q=0.05$)]{
				\label{Figure4-10}
				\centering
				\includegraphics[width=0.316\linewidth] {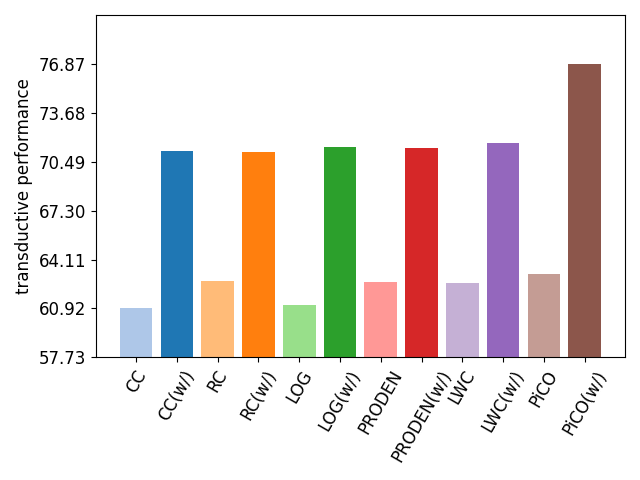}
			}
			\subfigure[Kuzushiju-MNIST ($q=0.5$)]{
				\label{Figure4-12}
				\centering
				\includegraphics[width=0.316\linewidth] {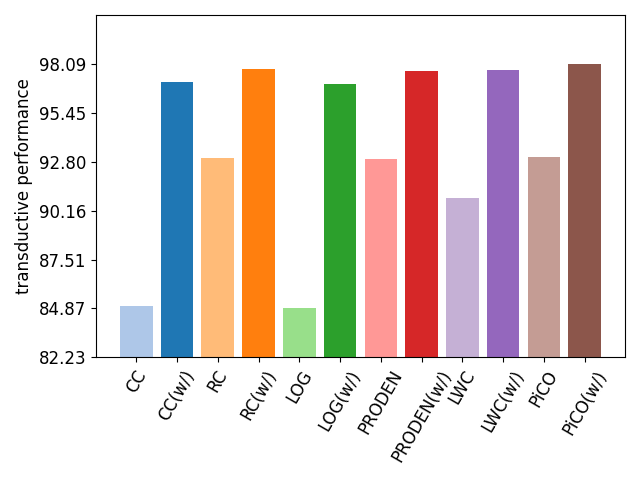}
			}
		\end{center}
		\caption{Transductive performance under different ambiguity levels. ``w/'' indicates the model with IRNet. The noise level is fixed to $\eta=0.3$.}
		\label{Figure4}
	\end{figure*}

	\subsection{Transductive Performance}
	Besides the inductive performance, we also evaluate the transductive performance of different approaches. Transductive results reflect the disambiguation ability on training data \cite{lyu2019gm, lv2020progressive}. In traditional PLL, the basic assumption is that the ground-truth label of each sample must be in the candidate set. To evaluate the transductive performance, previous works \cite{wang2022adaptive} generally determine the ground-truth label of each training sample as $\arg\max_{j \in S(x)}{f_j\left(x\right)}$. But this assumption is not satisfied in noisy PLL. Therefore, we predict the ground-truth label from the entire label space rather than the candidate set, i.e., $\arg\max_{j \in \mathcal{Y}}{f_j\left(x\right)}$. From Fig. \ref{Figure4}, we observe that IRNet always brings performance improvements to existing PLL methods. Compared with these baselines, IRNet utilizes the idea of label correction to purify noisy samples and reduce the noise level of the dataset, resulting in better transductive performance.

	\subsection{Noise Robustness}
	Last accuracy (i.e., test accuracy for the last epoch) reflects whether the model can prevent fitting noisy samples at the end of training. To reveal the noise robustness of our system, we further compare the last accuracy. Table \ref{Table4} shows the best accuracy, the last accuracy, and the performance gap between them. From this table, we observe that the system using IRNet can achieve similar best and last results in the presence of label noise. Taking the results on Kuzushiji-MNIST as an example, the performance gap of the system without IRNet is 9.42\%$\sim$12.46\%, while the performance gap of the system with IRNet is only 0.12\%$\sim$0.14\%. Therefore, with the help of IRNet, the system becomes stable and does not cause performance degradation due to overfitting when the number of epochs increases.

    \begin{table}[h]
		\centering
		\renewcommand\tabcolsep{5pt}
		\renewcommand\arraystretch{1.20}
		\caption{Test accuracy with and without IRNet under different ambiguity levels. The noise level of these datasets is fixed to $\eta=0.3$. We report the best accuracy, the last accuracy, and the performance gap between them.}
		\label{Table4}
		\begin{tabular}{c|c|ccc|ccc}
			\hline
			\multirow{2}{*}{Dataset} & \multirow{2}{*}{$q$} & \multicolumn{3}{c|}{with IRNet} & \multicolumn{3}{c}{without IRNet} \\
			& & Best & Last & $\Delta$ & Best & Last & $\Delta$ \\
			\hline \hline
			
			\multirow{3}{*}{CIFAR-10} & 0.1 &92.38&92.22&\textbf{0.16}&84.96&77.94&\textbf{7.02} \\
			& 0.3 &91.35&91.14&\textbf{0.21}&82.25&74.76&\textbf{7.49} \\
			& 0.5 &86.19&85.89&\textbf{0.30}&68.75&66.64&\textbf{2.11} \\
			\hline
			
			\multirow{3}{*}{CIFAR-100} & 0.01 &68.77&68.49&\textbf{0.28}&58.97&54.55&\textbf{4.42} \\
			& 0.03 &68.18&67.81&\textbf{0.37}&58.64&53.22&\textbf{5.42} \\
			& 0.05 &68.09&67.75&\textbf{0.34}&58.18&52.81&\textbf{5.37} \\
			\hline
			
			\multirow{3}{*}{\begin{tabular}[c]{@{}c@{}}Kuzushiji- \\ MNIST\end{tabular}} & 0.1 &97.95&97.81&\textbf{0.14}&96.53&87.11&\textbf{9.42} \\
			& 0.3 &97.87&97.75&\textbf{0.12}&95.34&85.06&\textbf{10.28} \\
			& 0.5 &95.84&95.71&\textbf{0.13}&90.87&78.41&\textbf{12.46} \\
			\hline
			
		\end{tabular}
	\end{table}

\begin{figure}[t]
		\begin{center}
			
			\subfigure[noise level on training data]{
				\label{Figure5-1}
				\centering
				\includegraphics[width=0.47\linewidth]{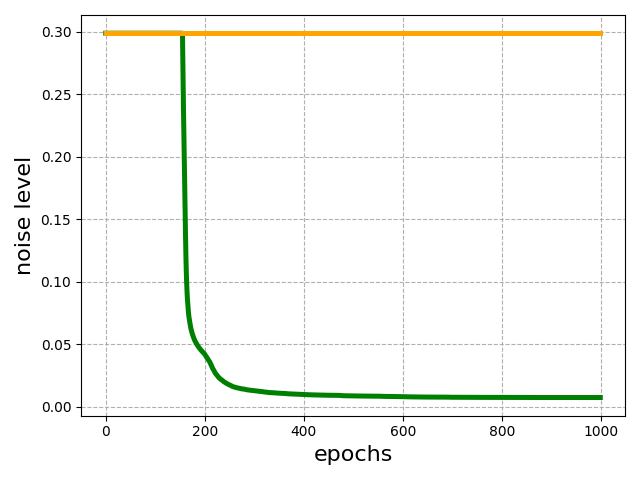}
			} 
			\subfigure[test accuracy]{
				\label{Figure5-2}
				\centering
				\includegraphics[width=0.47\linewidth]{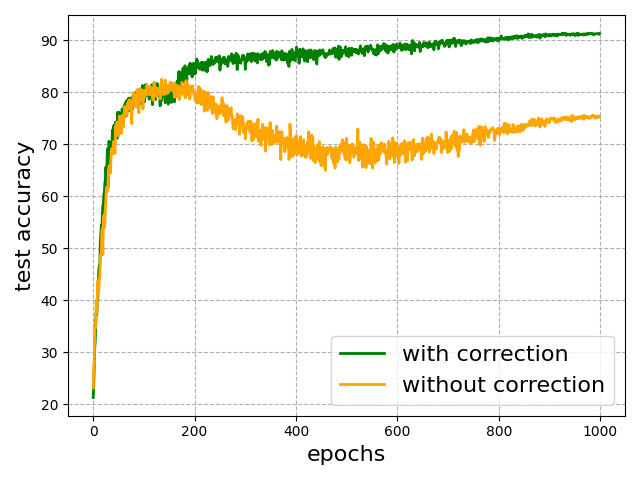}
			}
			
		\end{center}
		\caption{Performance comparison on CIFAR-10 ($q=0.3, \eta=0.3$) with and without IRNet correction. In these figures, we visualize the noise level on training data and test accuracy with increasing epochs.}
		\label{Figure5}
	\end{figure}

	Fig. \ref{Figure5} presents the curves of noise level and test accuracy. Without IRNet correction, the model fits noisy samples in the learning process and causes performance degradation on the test set. Differently, IRNet can reduce the noise level of the dataset (see Fig. \ref{Figure5-1}), thereby improving the test accuracy under noise conditions (see Fig. \ref{Figure5-2}). Meanwhile, based on theoretical analysis, the boundary $m$ should decrease exponentially in each round and eventually converge to a stable value $m_{\text{end}}$. Since $m$ is related to the noise level of the dataset $\eta$, $\eta$ should also decrease exponentially. Interestingly, we observe a similar phenomenon in Fig. \ref{Figure5-1}.

	\subsection{Importance of Smoothness Constraints}
	\label{sec:augmentation}
	IRNet relies on smoothness constraints to increase the reliability of prediction results. This process ensures the model's output is not significantly affected by plausible data augmentation. To reveal its importance, we compare the performance of different combinations of weak (i.e., SimAugment \cite{khosla2020supervised}) and strong (i.e., RandAugment \cite{cubuk2019randaugment}) augmentations. Experimental results are shown in Table \ref{Table5}.
	
	From Table \ref{Table5}, we observe that smoothness constraints can improve classification performance. Taking the results on CIFAR-10 as an example, the system with data augmentation outperforms the system without data augmentation by 0.50\%$\sim$1.90\%. With the help of smoothness constraints, we can leverage stricter criteria to select noisy samples, thereby increasing the reliability of predictions. These results verify the importance of smoothness constraints. 
	
	Meanwhile, we investigate the impact of different augmentation strategies. In Table \ref{Table5}, weak augmentation achieves better performance than strong augmentation. This phenomenon indicates that strong augmentation may not preserve all necessary information for classification. More notably, increasing the number of augmentations does not always bring better performance. More augmentations make the selection of noisy samples more stringent and the prediction results more credible. However, some noisy samples may not be corrected due to these strict rules. Therefore, there is a trade-off between them. Strict rules lead to some noisy samples not being corrected, while loose rules lead to some clean samples being erroneously corrected. We need to find a suitable augmentation strategy for IRNet.
	
	Table \ref{Table6} further compares the complexity of different strategies. Data augmentation requires additional feedforward processing on augmented samples, resulting in more computational cost and training time. Since feedforward processing does not change the model architecture, the number of parameters remains the same. Considering that increasing the number of augmentations does not always lead to better performance (see Table \ref{Table5}), we use one weak augmentation by default to speed up the training process.
	
	\begin{table}[t]
		\centering
		\renewcommand\arraystretch{1.20}
		\caption{Importance of smoothness constraints. In this table, \#W denotes the number of weak augmentations and \#S denotes the number of strong augmentations. We report test accuracy under varying ambiguity levels. The noise level of these datasets is fixed to $\eta=0.3$. The best performance is highlighted in bold.}
		\label{Table5}
		\begin{tabular}{c|cc|c|c|c}
			\hline
			Dataset & \#W & \#S & $q=0.1$ & $q=0.3$ & $q=0.5$ \\
			\hline
			\multirow{6}{*}{CIFAR-10} &0 & 0 &90.78 &89.87 &86.51 \\ 
			&0 & 1 &91.43 &90.30 &85.44 \\ 
			&1 & 0 &92.38 &91.35 &86.19 \\ 
			&0 & 2 &90.76 &89.32 &84.93 \\ 
			&1 & 1 &92.26 &91.32 &85.94 \\ 
			&2 & 0 &\textbf{92.68} &\textbf{91.49} &\textbf{87.01} \\ 
			\hline
			\hline
			
			Dataset & \#W & \#S & $q=0.01$ & $q=0.03$ & $q=0.05$ \\
			\hline
			\multirow{6}{*}{CIFAR-100} &0 & 0 &65.40 &63.93 &62.88 \\ 
			&0 & 1 &67.03 &65.63 &65.70 \\ 
			&1 & 0 &68.77 &\textbf{68.18} &\textbf{68.09} \\ 
			&0 & 2 &61.70 &59.58 &58.94 \\ 
			&1 & 1 &66.32 &65.13 &63.46 \\ 
			&2 & 0 &\textbf{69.44} &68.03 &66.90 \\ 
			\hline

		\end{tabular}
	\end{table}

	\begin{table}[t]
		\centering
		\renewcommand\tabcolsep{3.6pt}
		\renewcommand\arraystretch{1.20}
		\caption{Complexity comparison of different augmentation strategies on CIFAR-10. This table compares the MACs, number of parameters, training time per sample, and GPU memory usage.}
		\label{Table6}
		\begin{tabular}{c|c|c|c|c}
			\hline
			{\#W+\#S} & \begin{tabular}[c]{@{}c@{}}MACs \\ (G)\end{tabular} & \begin{tabular}[c]{@{}c@{}}Parameters \\ (M)\end{tabular} & \begin{tabular}[c]{@{}c@{}} Training Time \\ (ms)\end{tabular} & \begin{tabular}[c]{@{}c@{}}GPU Memory \\ (GB)\end{tabular} \\
			\hline
			\hline
			
			0 &1.11 &23.01 &1.08 &3.39 \\ 
			1 &1.67 &23.01 &1.18 &4.05 \\ 
			2 &2.23 &23.01 &1.25 &4.85 \\ 
			\hline
			
		\end{tabular}
	\end{table}

	\subsection{Parameter Sensitivity}
\begin{figure}[t]
		\begin{center}
			
			\subfigure[$\tau_\epsilon$ on CIFAR-10]{
				\label{Figure6-1}
				\centering
				\includegraphics[width=0.47\linewidth]{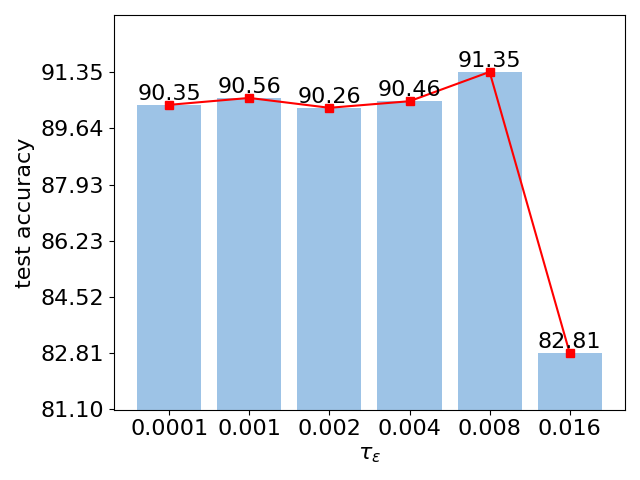}
			} 
			\subfigure[$e_0$ on CIFAR-10 ($\tau_\epsilon$=0.008)]{
				\label{Figure6-2}
				\centering
				\includegraphics[width=0.47\linewidth]{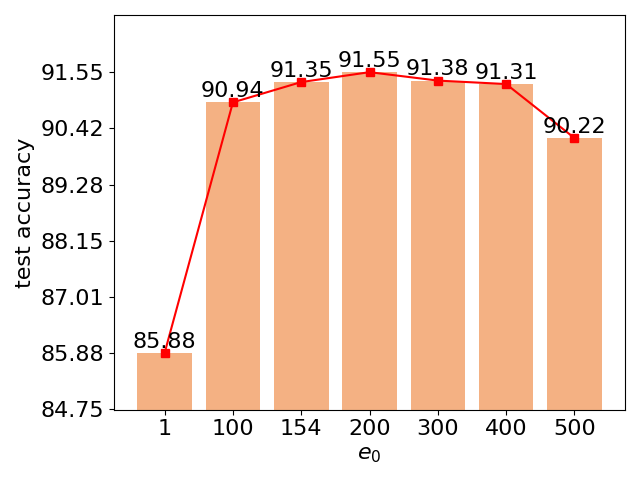}
			} 
			
			\subfigure[$\tau_\epsilon$ on CIFAR-100]{
				\label{Figure6-3}
				\centering
				\includegraphics[width=0.47\linewidth]{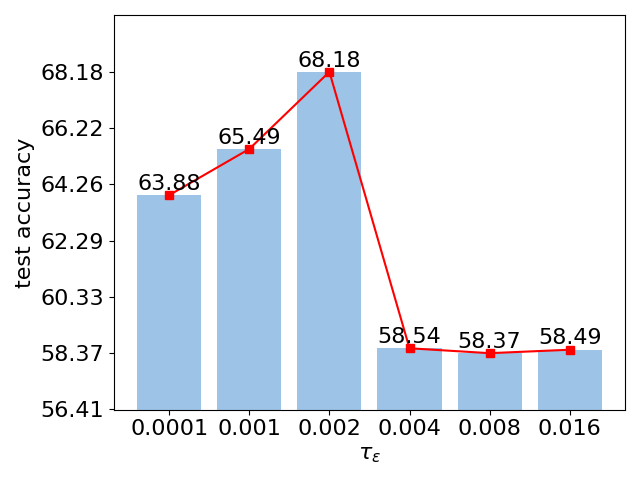}
			}
			\subfigure[$e_0$ on CIFAR-100 ($\tau_\epsilon$=0.002)]{
				\label{Figure6-4}
				\centering
				\includegraphics[width=0.47\linewidth]{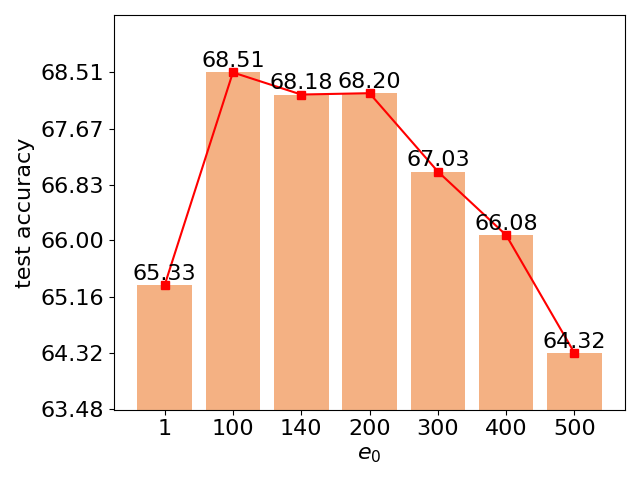}
			}
			
		\end{center}
		\caption{Parameter sensitivity analysis on CIFAR-10 ($q=0.3$) and CIFAR-100 ($q=0.03$). The noise level of these datasets is fixed to $\eta=0.3$.}
		\label{Figure6}
	\end{figure}

	IRNet mainly contains two hyperparameters, i.e., the correction epoch $e_0$ and the margin $\tau_\epsilon$. We automatically determine $e_0$ and manually select $\tau_\epsilon$. In this section, we conduct a parameter sensitivity analysis to reveal the impact of these hyperparameters. Experimental results are shown in Fig. \ref{Figure6}.
	
	In Fig. \ref{Figure6-1} and Fig. \ref{Figure6-3}, IRNet performs poorly when $\tau_\epsilon$ is too large or too small. A large $\tau_\epsilon$ makes the selection of noisy samples strict but causes some noisy samples not to be corrected. A small $\tau_\epsilon$ reduces the requirement for noisy samples but causes some clean samples to be erroneously corrected. Therefore, a suitable $\tau_\epsilon$ can improve classification performance under noisy conditions.
	
	Then, we analyze the influence of $e_0$ under the optimal $\tau_\epsilon$. In Fig. \ref{Figure6-2} and Fig. \ref{Figure6-4}, the automatically determined $e_0$ is 154 and 140, respectively. We also compare with $e_0$ in $\{1, 100, 200, 300, 400, 500\}$. The performance of IRNet depends on two key modules: noisy sample detection and label correction. Since these modules perform poorly at the early or late epochs, the classification performance first increases and then decreases with increasing $e_0$. More notably, our automatic selection strategy can achieve competitive performance with less manual effort. These results verify the effectiveness of our selection strategy. Meanwhile, the suitable range of $e_0$ is relatively large. Taking the results on CIFAR-10 as an example, the test accuracy is very close when $e_0$ ranges from 100 to 400.

	\begin{figure*}[t]
		\begin{center}
			\subfigure[LOG]{
				\label{Figure7-1}
				\centering
				\includegraphics[width=0.226\linewidth, trim=36 36 36 36] {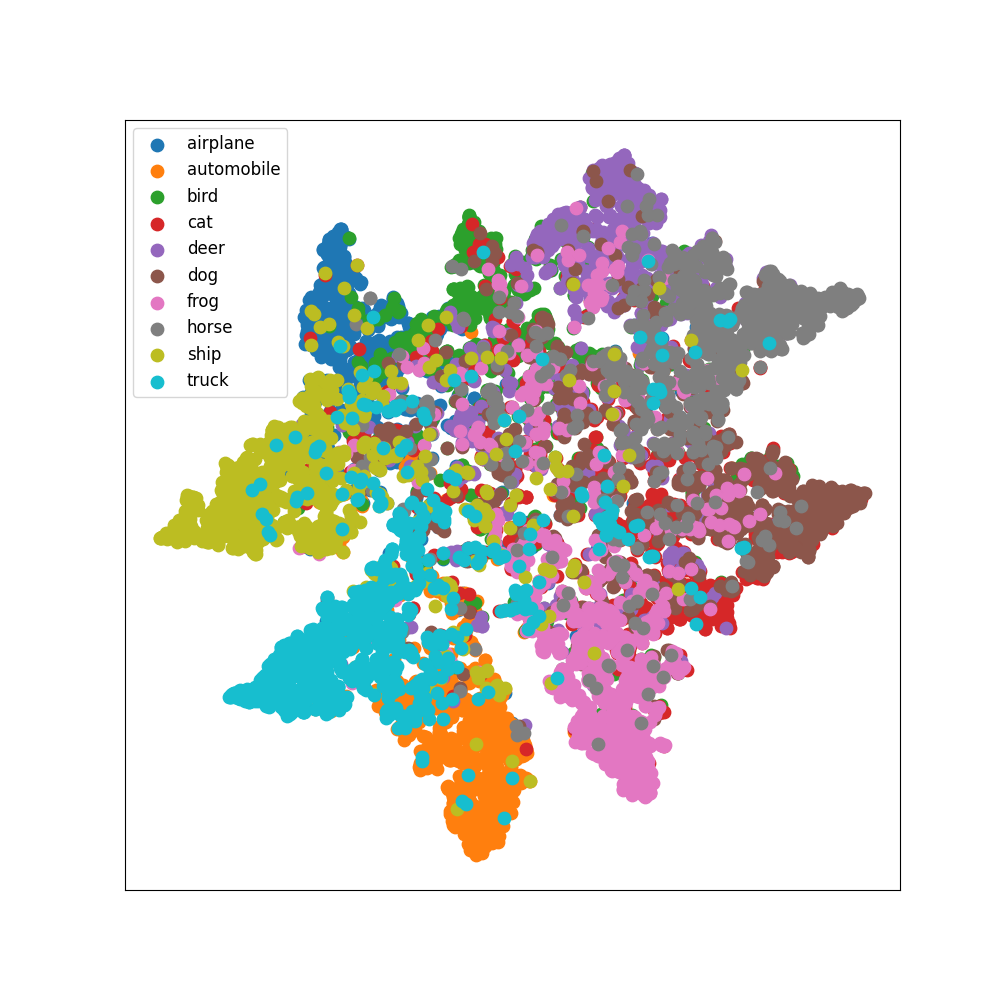}
			}
			\subfigure[RC]{
				\label{Figure7-2}
				\centering
				\includegraphics[width=0.226\linewidth, trim=36 36 36 36]	{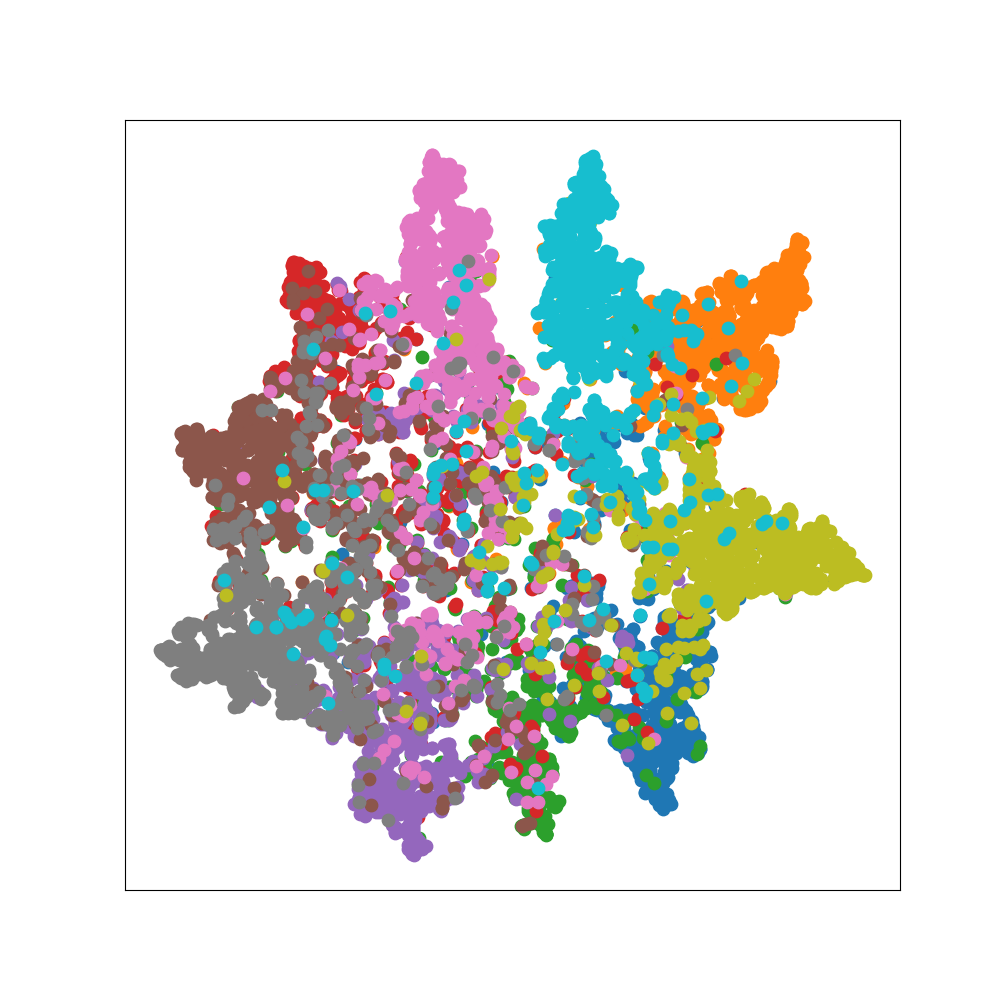}
			}
			\subfigure[PiCO]{
				\label{Figure7-3}
				\centering
				\includegraphics[width=0.226\linewidth, trim=36 36 36 36] {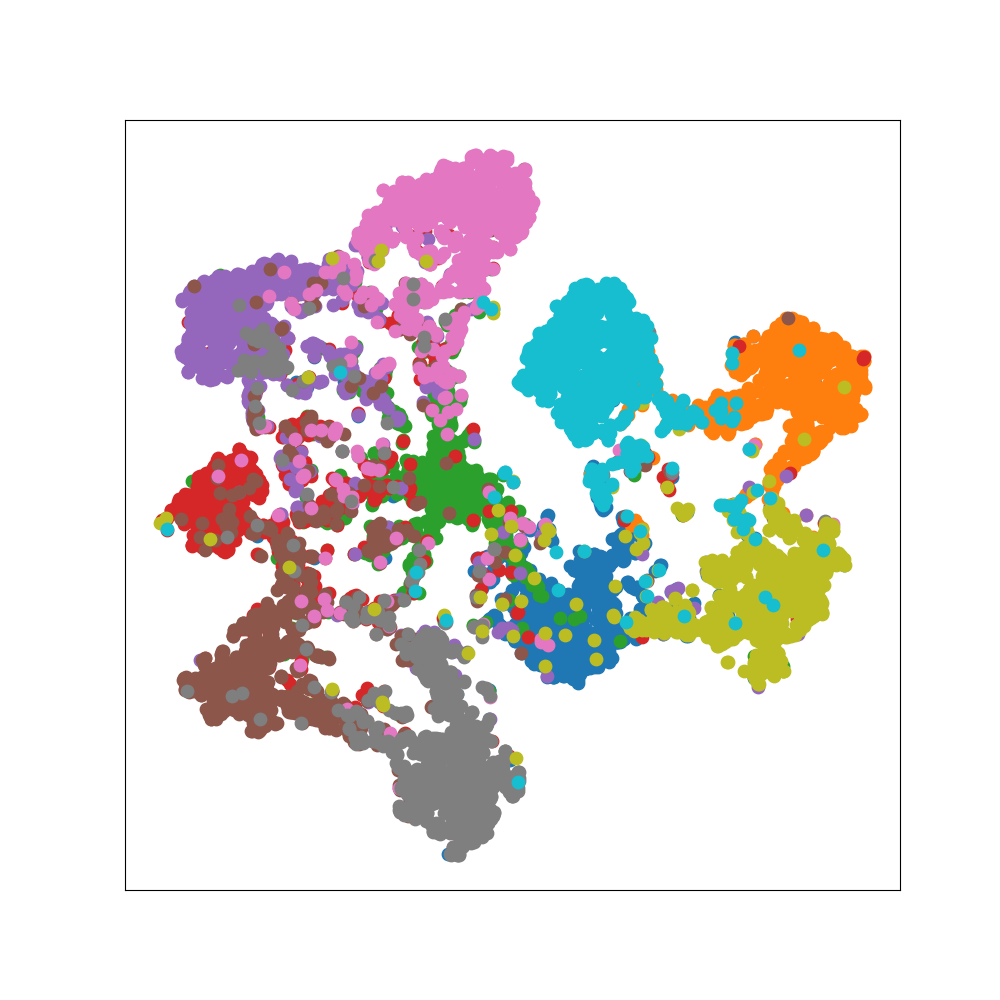}
			}
			\subfigure[MSE]{
				\label{Figure7-4}
				\centering
				\includegraphics[width=0.226\linewidth, trim=36 36 36 36] {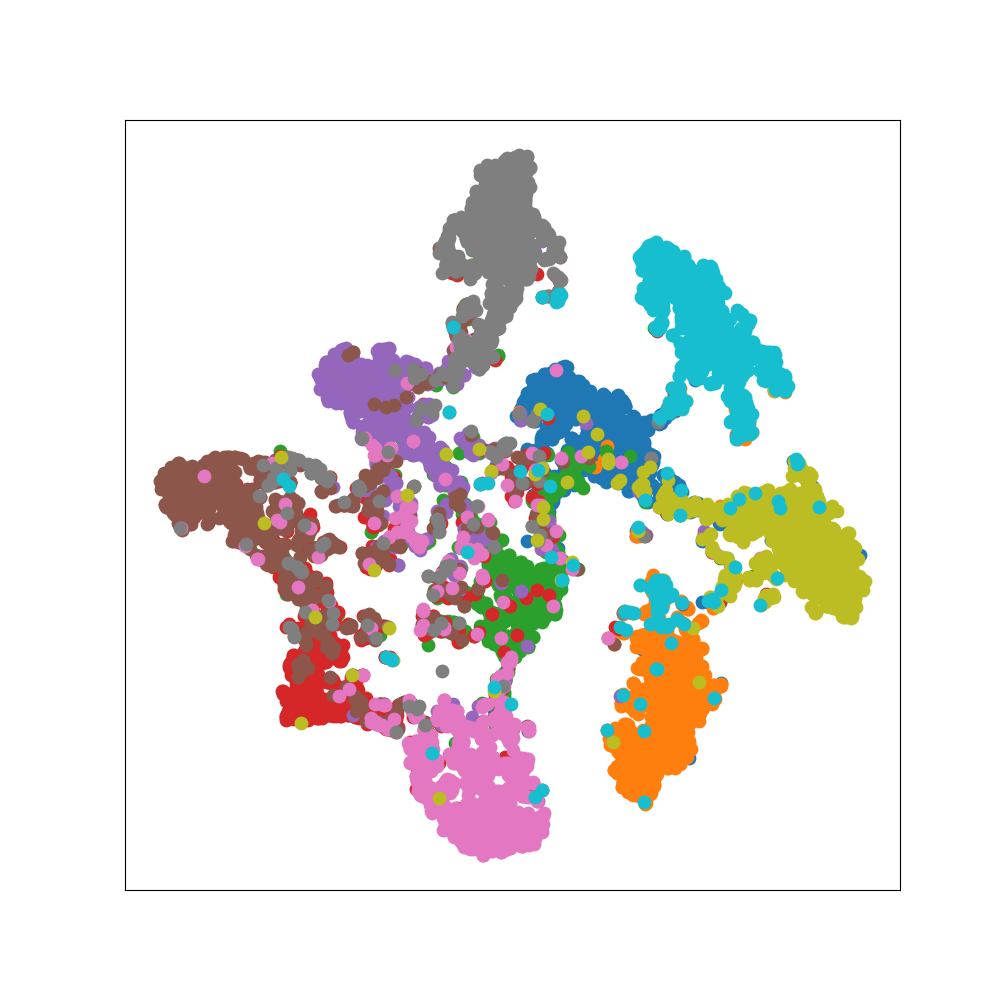}
			}
			\subfigure[SCE]{
				\label{Figure7-5}
				\centering
				\includegraphics[width=0.226\linewidth, trim=36 36 36 36] {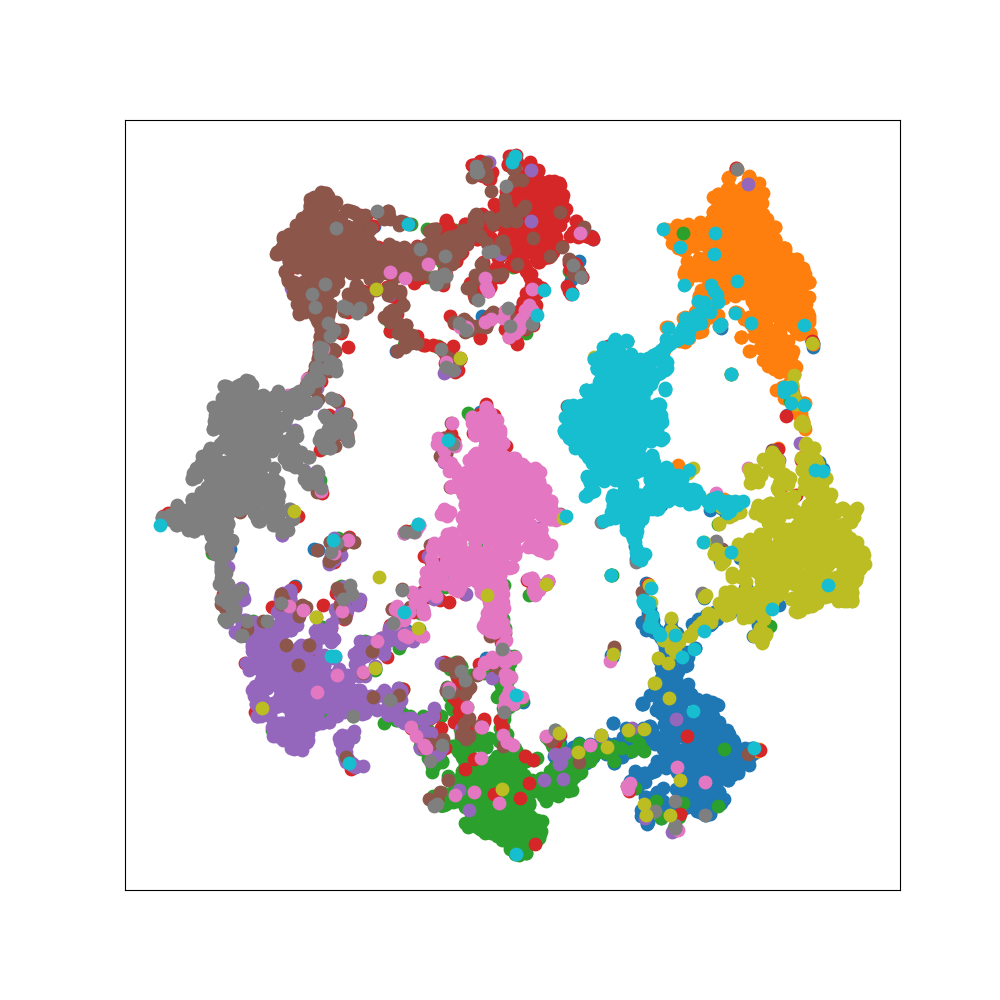}
			}
			\subfigure[GCE]{
				\label{Figure7-6}
				\centering
				\includegraphics[width=0.226\linewidth, trim=36 36 36 36] {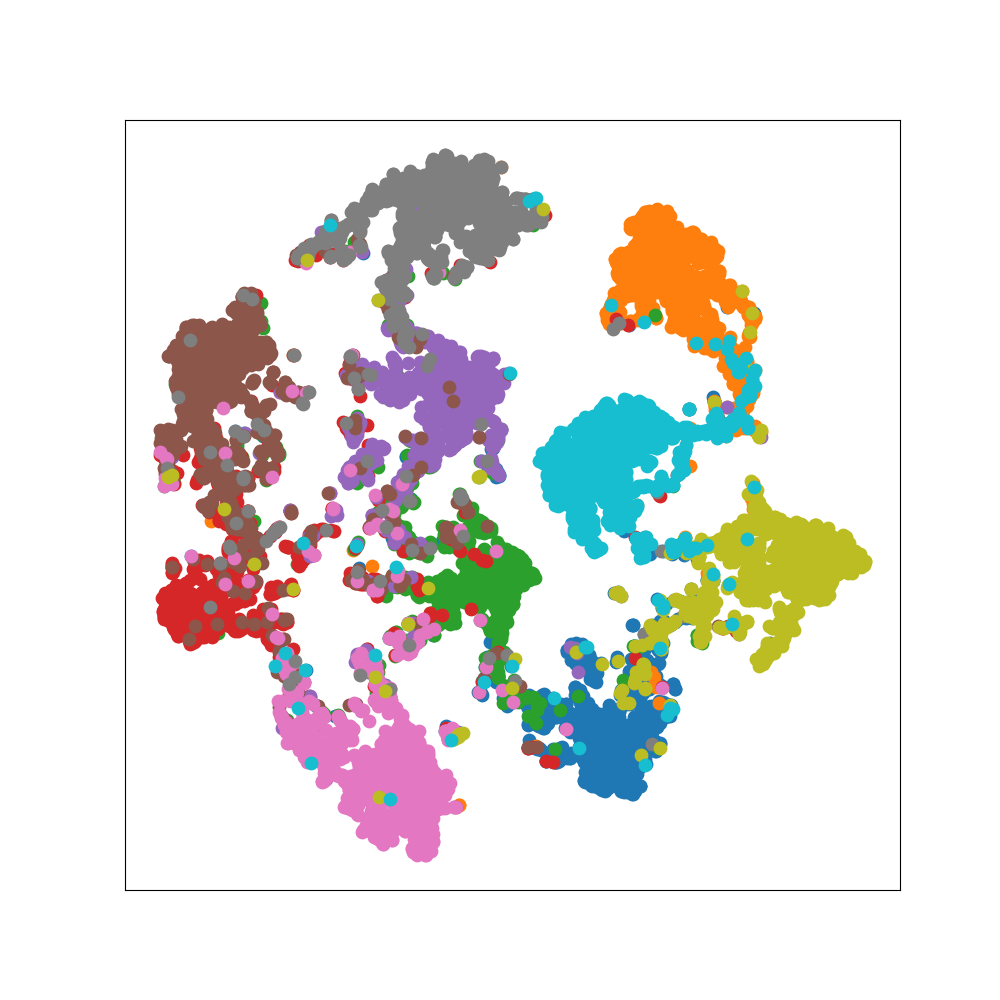}
			}
			\subfigure[MAE]{
				\label{Figure7-7}
				\centering
				\includegraphics[width=0.226\linewidth, trim=36 36 36 36] {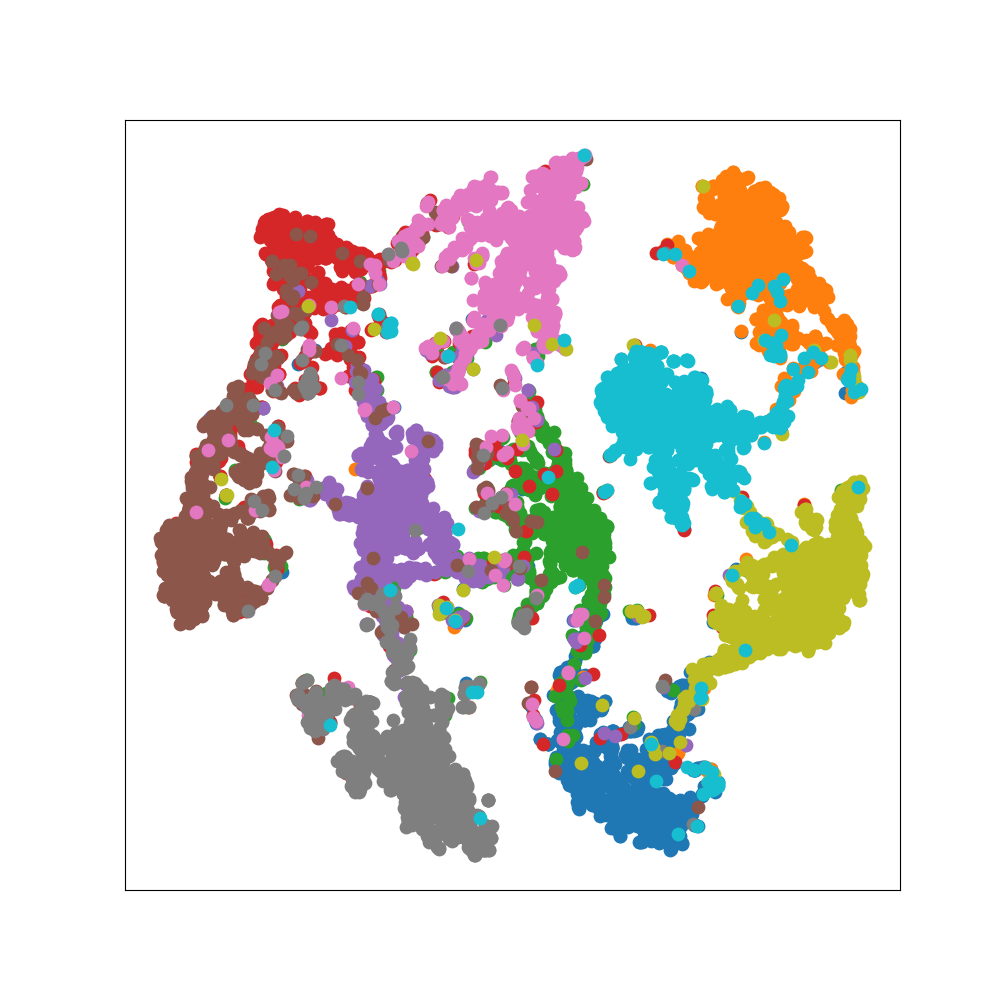}
			}
			\subfigure[IRNet]{
				\label{Figure7-8}
				\centering
				\includegraphics[width=0.226\linewidth, trim=36 36 36 36] {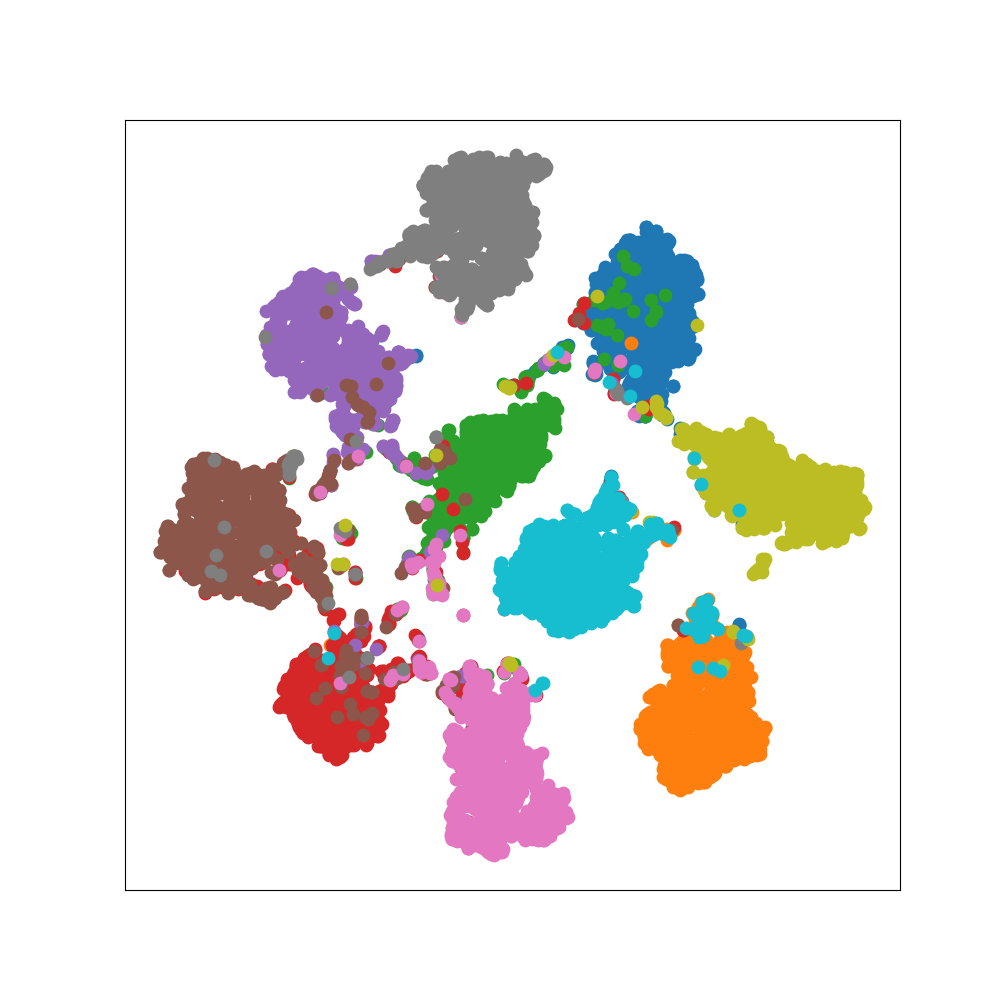}
			}
		\end{center}
		\caption{T-SNE visualization results of different methods on the CIFAR-10 test set ($q=0.3, \eta=0.3$).}
		\label{Figure7}
	\end{figure*} 
	
	\begin{figure*}[t]
		\begin{center}
			\subfigure[100 epochs]{
				\label{Figure8-1}
				\centering
				\includegraphics[width=0.226\linewidth, trim=36 36 36 36] {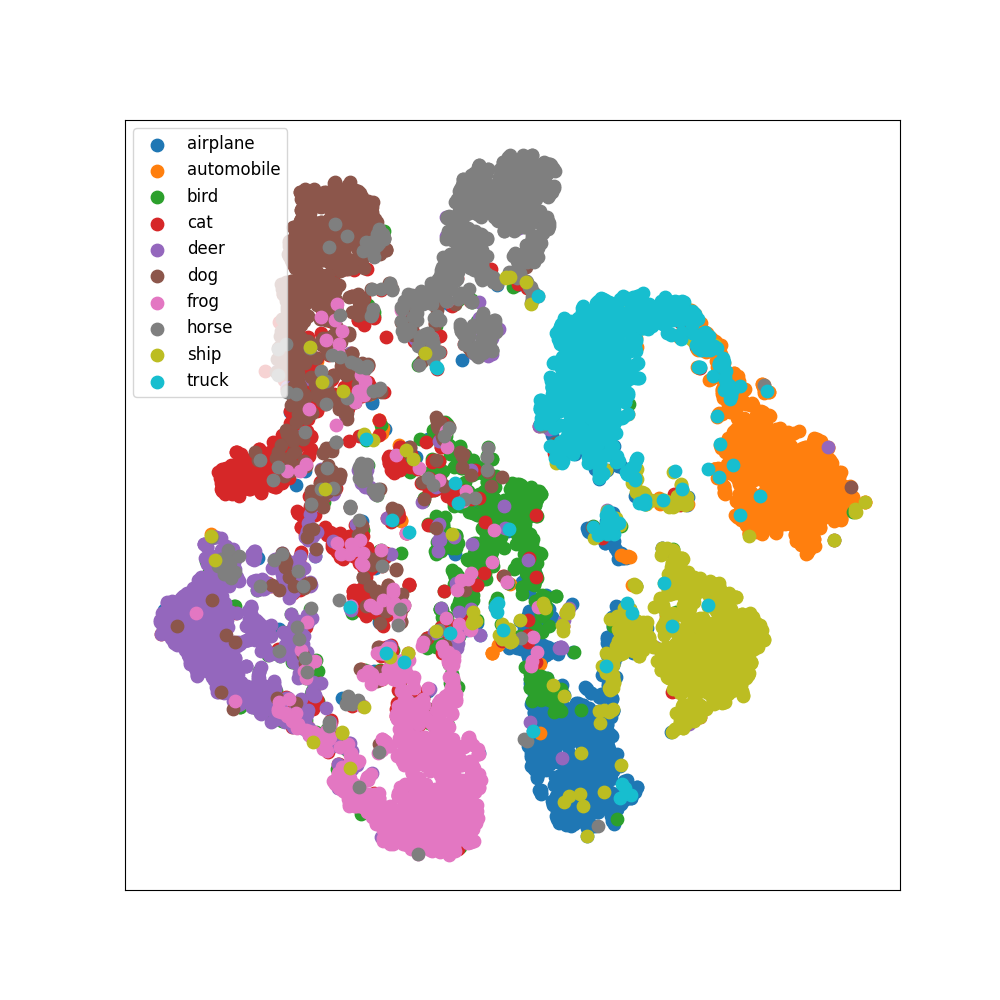}
			}
			\subfigure[200 epochs]{
				\label{Figure8-2}
				\centering
				\includegraphics[width=0.226\linewidth, trim=36 36 36 36] {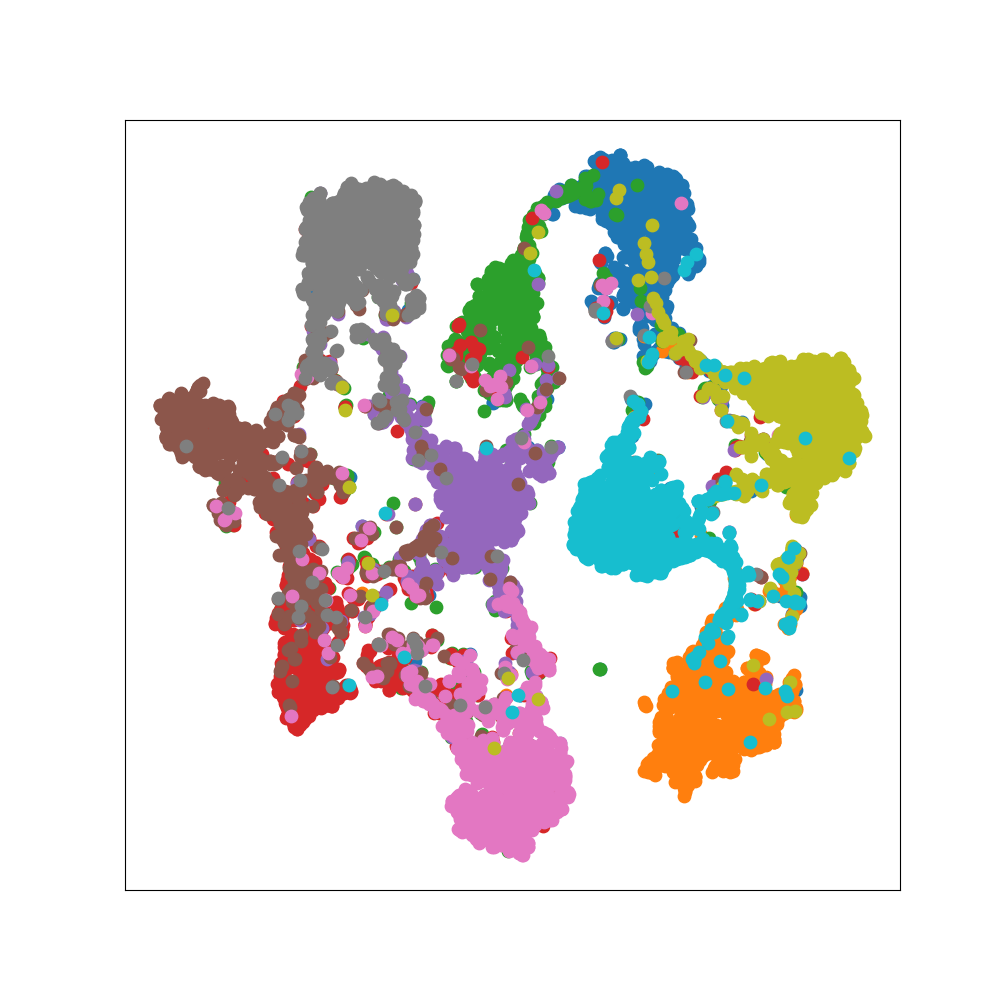}
			}
			\subfigure[500 epochs]{
				\label{Figure8-3}
				\centering
				\includegraphics[width=0.226\linewidth, trim=36 36 36 36] {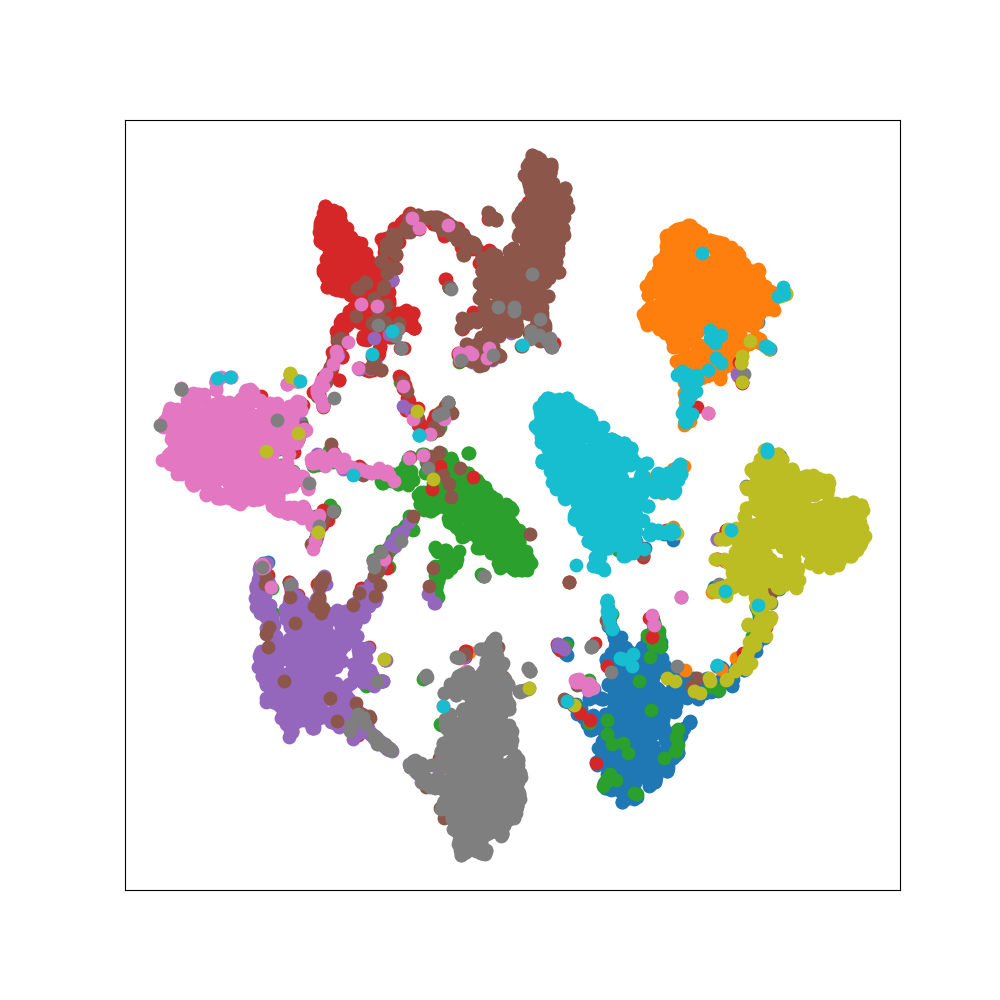}
			}
			\subfigure[910 epochs]{
				\label{Figure8-4}
				\centering
				\includegraphics[width=0.226\linewidth, trim=36 36 36 36] {image/show_arnet_30_910}
			}
		\end{center}
		\caption{T-SNE visualization results of IRNet on the CIFAR-10 test set ($q=0.3, \eta=0.3$) with increasing training epochs.}
		\label{Figure8}
	\end{figure*} 
	
	\subsection{Role of Swapping}
	\label{sec:role_of_swapping}
	In this paper, we correct the noisy sample by moving the predicted target into the candidate set. This process leads to an increase in the number of candidate labels $|S(x)|$. Previous works have demonstrated that large $|S(x)|$ may increase the difficulty of model optimization, resulting in a decrease in classification performance \cite{feng2020provably}. To keep $|S(x)|$ consistent, a heuristic solution is to further move out the candidate label with the lowest confidence (i.e., $\bar{y}=\arg\min_{j \in S(x)}{f_j\left(x\right)}$). To sum up, for each noisy sample, we move the predicted target into $S(x)$ and move $\bar{y}$ out of $S(x)$. We call this strategy \emph{swapping}. To investigate its role in noisy PLL, we conduct additional experiments in Table \ref{Table7}. The system without swapping is identical to IRNet.

	\begin{table}[h]
		\centering
		\renewcommand\arraystretch{1.20}
		\caption{Test accuracy with and without swapping under varying ambiguity levels. The noise level of these datasets is fixed to $\eta=0.3$.}
		\label{Table7}
		\begin{tabular}{c|c|c|c}
			\hline
			Dataset & $q$ &with swapping & without swapping \\
			\hline \hline
			
			\multirow{3}{*}{CIFAR-10} & 0.1 &91.48 &\textbf{92.38} \\
			& 0.3 &90.65 &\textbf{91.35} \\
			& 0.5 &84.15 &\textbf{86.19} \\
			\hline
			
			\multirow{3}{*}{CIFAR-100} & 0.01 &67.62 &\textbf{68.77} \\
			& 0.03 &67.57 &\textbf{68.18} \\
			& 0.05 &67.19 &\textbf{68.09} \\
			\hline
			
		\end{tabular}
	\end{table}
	
	In Table \ref{Table7}, the swapping operation does not improve the classification performance. Taking the results on CIFAR-10 as an example, the system with swapping performs slightly worse than the system without swapping by 0.70\%$\sim$2.04\%. For noisy samples, the swapping operation can keep the number of candidate labels consistent by moving $\bar{y}$ out of $S(x)$. However, the prediction result $f\left(x\right)$ is not always correct. Due to wrong predictions, it may also move the ground-truth label out of $S(x)$. 

	\subsection{Visualization of Embedding Space}
	Besides quantitative results, we also conduct qualitative analysis to show the strengths of our method. We exploit t-distributed stochastic neighbor embedding (t-SNE) \cite{van2008visualizing}, a tool widely utilized for high-dimensional data visualization. Fig. \ref{Figure7} presents the visualization results of different methods. From these figures, we observe that our IRNet can effectively disentangle the latent representation and reveal the underlying class distribution. In Fig. \ref{Figure8}, we further visualize the latent features of IRNet with increasing training epochs. From these figures, we observe that the separation among classes becomes clear during training. These results also show the effectiveness of our method for noisy PLL.

	\subsection{Practical Implications of Noisy PLL}
        \label{sec:practical_implications}
	To verify the necessity of noisy PLL in real-world scenarios, we compare the performance of different learning algorithms on RAF-DB, a crowdsourced facial expression dataset (see Section \ref{Sec4-1} for more details). For supervised learning and noisy label learning, we treat the dominant emotion as the ground-truth label. For PLL and noisy PLL, we use crowdsourced annotations to build the candidate set. We choose typical noisy label learning, PLL, and noisy PLL algorithms for comparison, which are DivideMix, PiCO, and our IRNet, respectively. For a fair comparison, all methods use the same backbone and data augmentation strategy.
	
	In Table \ref{Table8}, we observe that PLL and noisy PLL achieve better performance than supervised learning and noisy label learning. Due to the complexity of emotions, the dominant label generally contains errors, but the candidate set is more comprehensive. Meanwhile, in Table \ref{Table8}, the performance of noisy PLL is better than that of PLL. The candidate set generated by crowdsourcing does not always cover the ground-truth label. With the help of IRNet, we can purify the candidate set and achieve better performance. Therefore, the noisy PLL method can be considered complementary to current learning algorithms.
	
	\begin{table}[t]
		\centering
		\renewcommand\arraystretch{1.20}
		\caption{Performance comparison of different learning algorithms on RAF-DB.}
		\label{Table8}
		\begin{tabular}{c|c}
			\hline
			Method & Accuracy\\
			\hline
			\hline
			supervised learning & 64.44 \\
			noisy label learning (DivideMix) & 69.37 \\
			partial label learning (PiCO) & 72.54 \\
			noisy partial label learning (IRNet)	 & 77.15 \\
			\hline
		\end{tabular}
	\end{table}

    \section{Discussion}
	\label{sec7}
	Noisy PLL is fundamentally distinct from Learning from Noisy Labels (LNL) in its task definition, and distinct task formulations inherently yield different solutions.
    
    \textbf{(1) Task Definition.} \emph{LNL aims to mitigate overfitting on mislabeled samples.} Specifically, let $\mathcal{Y}=\left\{1, 2, \cdots, C\right\}$ denote the label space with $C$ distinct classes. We consider a dataset $\{(x_{i}, y_{i})\}_{i=1}^{N}$ consisting of $N$ samples. In supervised learning, the label $y_i$ is assumed to be correct; in LNL, this assumption is relaxed, allowing some samples to have incorrect labels. \emph{The task definition of noisy PLL differs fundamentally from that of LNL.} Specifically, we consider a partially labeled dataset $\{(x_{i}, S(x_{i})\}_{i=1}^{N}$, where the function $S(\cdot)$ maps each sample $x_i$ to its corresponding candidate set $S(x_i) \subseteq \mathcal{Y}$. In PLL, the ground-truth label $y_i$ must be concealed within the candidate set, i.e., $y_i \in S(x_i)$. In noisy PLL, this assumption is relaxed, allowing some samples whose ground-truth label is not concealed within the candidate set.
    
    \textbf{(2) Task Necessity.} In Section \ref{sec:practical_implications}, experimental results show that the noisy PLL approach outperforms LNL and PLL in real-world applications, further validating the necessity of the noisy PLL task.

    \textbf{(3) Solutions.} Several works have also focused on noisy PLL. For instance, Lv et al. \cite{lv2023robustness} employed noise-robust loss functions to address noisy PLL. Wang et al. \cite{wang2023pico+} proposed a distance-based clean sample selection method and learned robust classifiers in a semi-supervised learning framework. However, in Table \ref{Table3}, our IRNet outperforms these approaches under noisy conditions. This highlights that different solutions for noisy PLL yield distinct performance outcomes, and this task cannot be effectively addressed by directly applying solutions from LNL.
    
    In summary, noisy PLLs are fundamentally different from LNLs in terms of task definition, demonstrate practicality in real-world applications, and require carefully designed solutions. These conclusions stress the importance of further discussion and research on noisy PLL.

	\section{Conclusions}
	\label{sec6}
	This paper extends the traditional PLL problem to the more challenging noisy PLL problem, allowing some samples whose ground-truth labels are not in their candidate sets. To address this task, we propose IRNet, a plug-in framework with strong motivation and theoretical guarantees. It can be easily integrated with existing PLL methods, achieving better inductive and transductive performance under noisy conditions. Meanwhile, we verify the noise robustness of our method and reveal the importance of smoothness constraints. We also perform parameter sensitivity analysis and present the impact of different hyperparameters.
	
	In this paper, we focus on the scenario where both clean and noisy samples have their ground-truth labels in a predefined label space. In the future, we will explore some challenging scenarios. For example, there are out-of-distribution samples in the dataset, and the ground-truth labels of these samples are outside the label space. Meanwhile, this paper requires manual tuning of $\tau_\epsilon$. In our future work, we aim to explore automatic selection methods for $\tau_\epsilon$, making IRNet entirely parameter-free.

	\section*{Acknowledgements}
	This work is supported by the Excellent Youth Program of State Key Laboratory of Multimodal Artificial Intelligence Systems (MAIS2024311) and the National Natural Science Foundation of China (NSFC) (No.62201572, No.61831022, No.62276259, No.U21B2010).

	\bibliographystyle{IEEEtran}
	\bibliography{mybib}

	\begin{IEEEbiography}[{\includegraphics[width=1.1in,height=1.25in,clip,keepaspectratio]{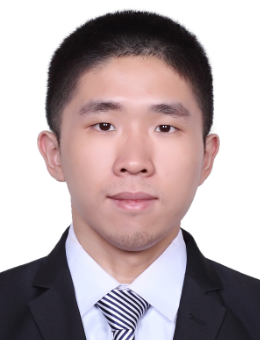}}]{Zheng Lian}
		(IEEE Senior Member) is an Associate Professor at the State Key Laboratory of Multimodal Artificial Intelligence Systems, Institute of Automation, Chinese Academy of Sciences (CASIA). He received his Ph.D degree from the CASIA in 2021. His research interest primarily centers on human-centric AI, affective computing, and large language models. He co-organized a series of challenges and workshops (MER@IJCAI, MRAC@ACM Multimedia, MEIJU@ICASSP). He (co-)authored more than 90 publications in journals, patents, and conference proceedings, leading to $>$3,500 citations (h-index: 33). He has published multiple papers on TPAMI, ICML, NeurlPS, TNNLS, TAC, TASLP, etc. He also serves as Associate Editor of IEEE Transactions on Affective Computing and IEEE Transactions on Audio, Speech, and Language Processing, Area Editor of Information Fusion, Area Chair of ACM Multimedia, and Conference Program Committee of NeurIPS, ICLR, ICML, etc.
	\end{IEEEbiography}
	\begin{IEEEbiography}[{\includegraphics[width=1.1in,height=1.25in,clip,keepaspectratio]{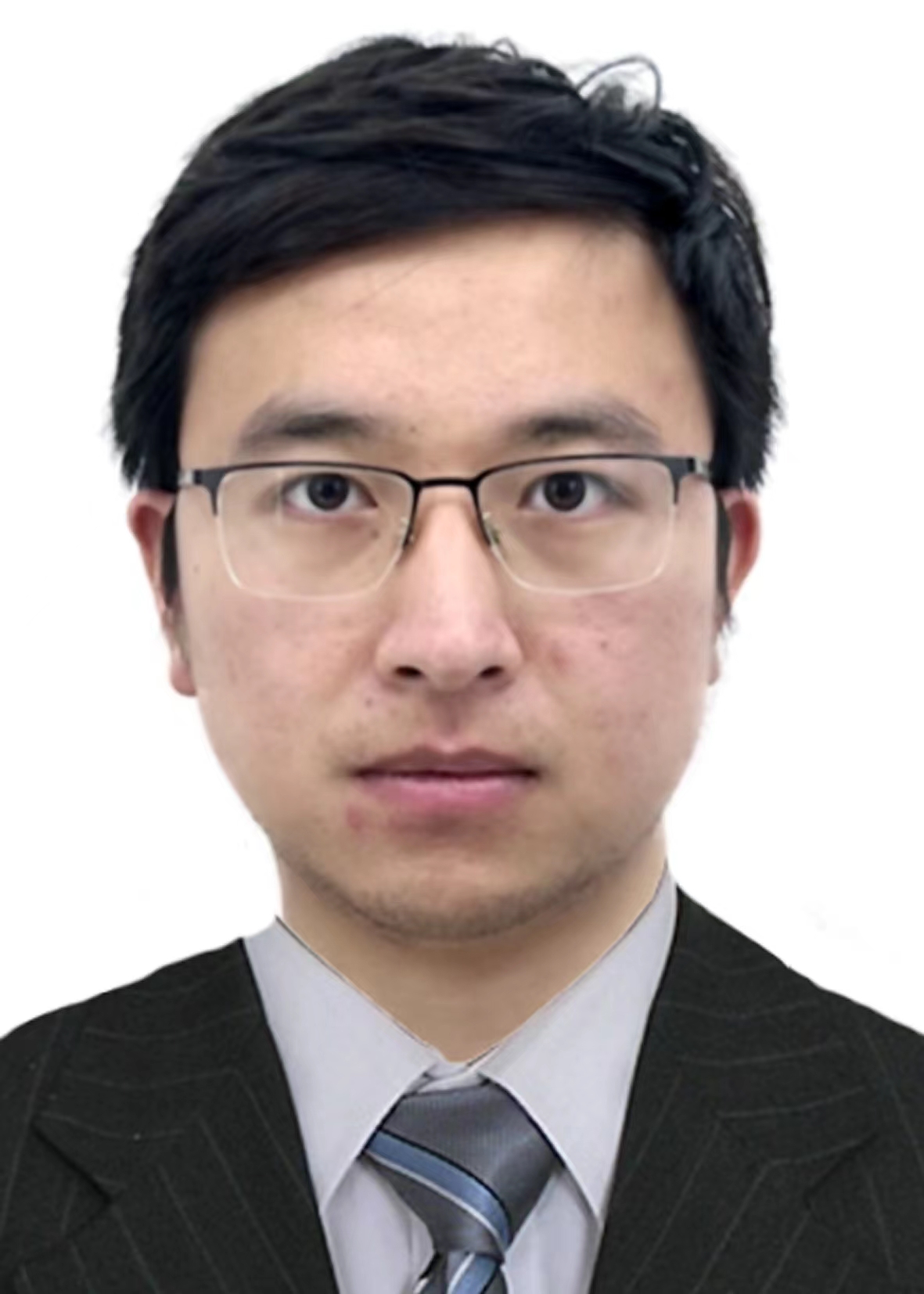}}]{Mingyu Xu}
		received the B.S. degree from Peking University, Beijing, China, in 2021 and the M.S. degree in 2024 from the Institute of Automation, China Academy of Sciences, Beijing, China. He is now engaged in artificial intelligence research in the seed group of ByteDance, and his current research interests include machine learning and large language modeling. He has published multiple papers on NeurIPS and ICML and served as a reviewer.
	\end{IEEEbiography}
	\begin{IEEEbiography}[{\includegraphics[width=1.1in,height=1.25in,clip,keepaspectratio]{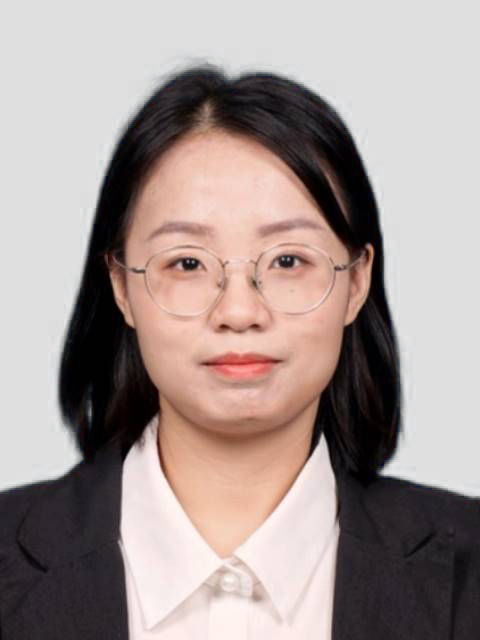}}]{Lan Chen}
		received the B.S. degree from the China University of Petroleum, Beijing, China, in 2016. And she received the Ph.D degree from the Institute of Automation, Chinese Academy of Sciences, Beijing, China, in 2022. Her current research interests include noisy label learning and image processing.
	\end{IEEEbiography}
	\begin{IEEEbiography}[{\includegraphics[width=1.1in,height=1.25in,clip,keepaspectratio]{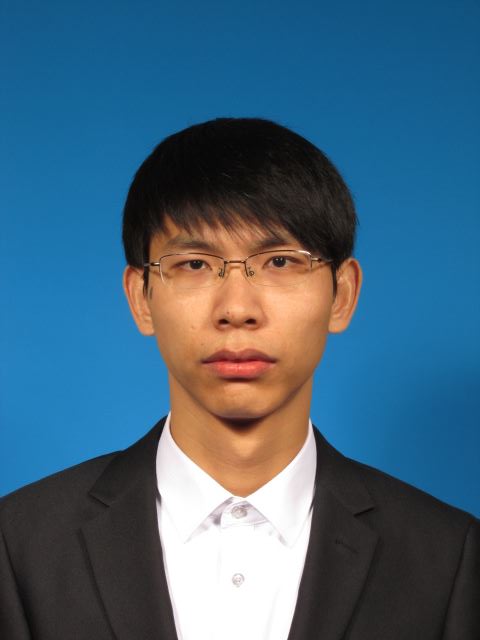}}]{Licai Sun}
        earned his Ph.D. from the Institute of Automation, Chinese Academy of Sciences (CASIA), and the University of Chinese Academy of Sciences (UCAS). He is currently a postdoctoral researcher working alongside Academy Professor Guoying Zhao at the University of Oulu and Professor Jukka Leppänen at the University of Turku. His research interests lies in the intersection of affective computing and deep learning, with a particular emphasis on multimodal learning and self-supervised learning.
	\end{IEEEbiography}
	\begin{IEEEbiography}[{\includegraphics[width=1in,height=1.25in,clip]{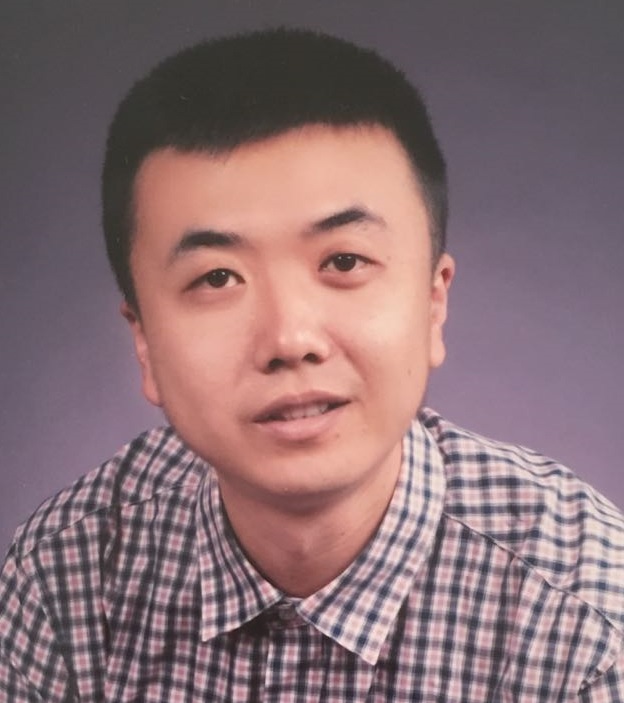}}]{Bin Liu}
		received the Ph.D. degree from the Institute of Automation, Chinese Academy of Sciences, Beijing, China, in 2015. He is currently an Associate Professor in the Institute of Automation, Chinese Academy of Sciences, Beijing, China. His current research interests include affective computing and audio signal processing.
	\end{IEEEbiography}
	\begin{IEEEbiography}[{\includegraphics[width=1.1in,height=1.3in,clip,keepaspectratio]{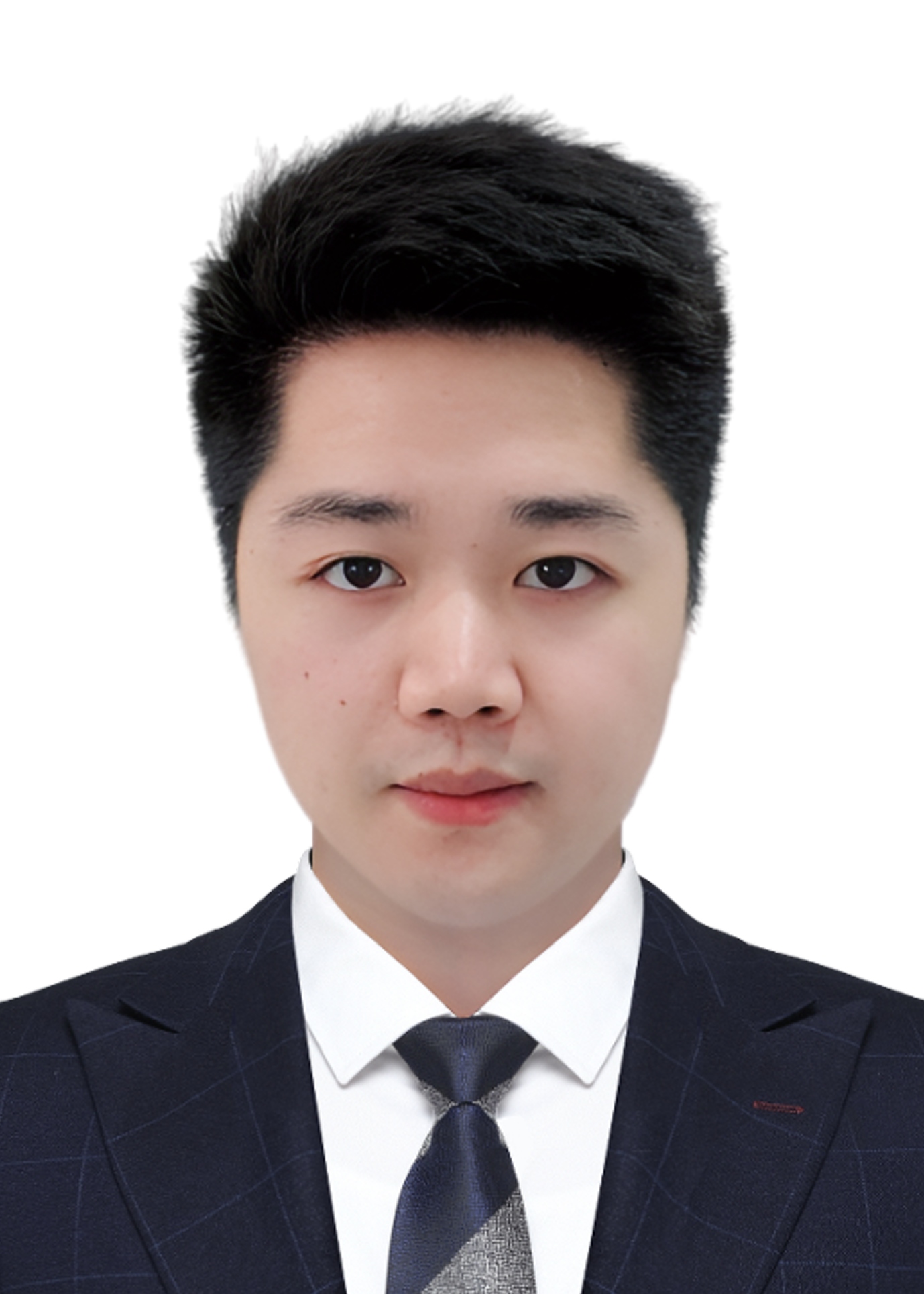}}]{Lei Feng} 
		received his Ph.D. degree from Nanyang Technological University, Singapore, in 2021. He is currently a Full Professor at the School of Computer Science and Engineering, Southeast University, China. His main research interests include trustworthy deep learning and multimodal large language models. He has published more than 100 papers at premier conferences and journals. He has regularly served as an Area Chair for ICML, NeurIPS, and ICLR, and served as an Action Editor for Neural Networks and Transactions on Machine Learning Research. He has received the ICLR 2022 Outstanding Paper Award Honorable Mention and the WAIC 2024 Youth Outstanding Paper Nomination Award. He was named to Forbes 30 Under 30 China 2021, Forbes 30 Under 30 Asia 2022, and Asia-Pacific Leaders Under 30 in 2024.
	\end{IEEEbiography}
	\begin{IEEEbiography}[{\includegraphics[width=1.1in,height=1.25in,clip,keepaspectratio]{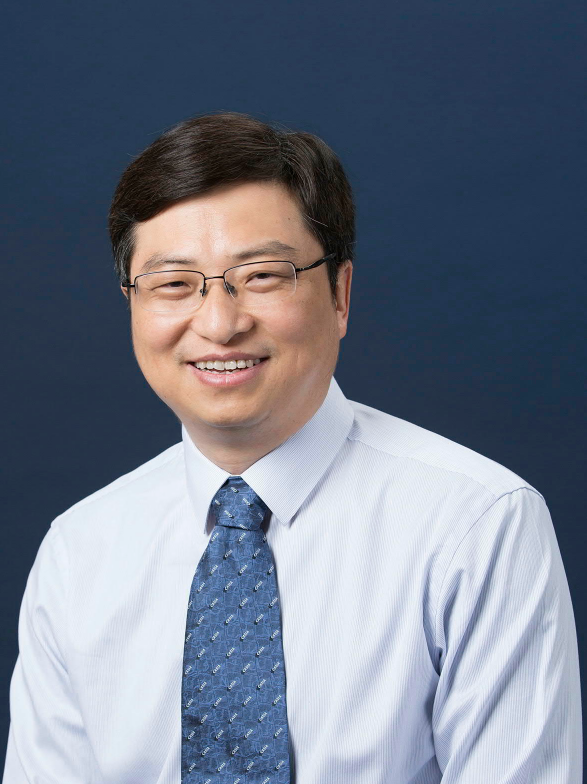}}]{Jianhua Tao}
		(IEEE Senior Member) received the Ph.D. degree from Tsinghua University, Beijing, China, in 2001, and the M.S. degree from Nanjing University, Nanjing, China, in 1996. He is currently a Professor with the Department of Automation, Tsinghua University, Beijing, China. He has authored or coauthored more than eighty papers in major journals and proceedings. His current research interests include speech recognition, speech synthesis and coding methods, human–computer interaction, multimedia information processing, and pattern recognition. He is the Chair or Program Committee Member for several major conferences, including ICPR, ACII, ICMI, ISCSLP, etc. He is also the Steering Committee Member for the IEEE Transactions on Affective Computing, an Associate Editor for the Journal on Multimodal User Interface and the International Journal on Synthetic Emotions, and the Deputy Editor-in-Chief for the Chinese Journal of Phonetics.
	\end{IEEEbiography}
	
\end{document}